\documentclass{article}
\PassOptionsToPackage{numbers, sort&compress}{natbib}
\usepackage[final]{neurips_2023}
\usepackage{wrapfig}
\usepackage[utf8]{inputenc} % allow utf-8 input
\usepackage[T1]{fontenc}    % use 8-bit T1 fonts
\usepackage{hyperref}       % hyperlinks
\usepackage{url}            % simple URL typesetting
\usepackage{booktabs}       % professional-quality tables
\usepackage{amsfonts}       % blackboard math symbols
\usepackage{nicefrac}       % compact symbols for 1/2, etc.
\usepackage{microtype}      % microtypography

\usepackage{algorithm}
\usepackage{algorithmic}
\usepackage{float}
\usepackage{macros}
\usepackage{neurips_2023}
\usepackage{caption}
\usepackage{subcaption}
\usepackage[rightcaption]{sidecap}
\sidecaptionvpos{figure}{c}
\usepackage{graphicx}

\usepackage{natbib}
\newcommand{\cv}[1]{\textcolor{black}{#1}}

\title{Anchor Data Augmentation}

\author{%
Nora Schneider$^{1}$ \quad Shirin Goshtasbpour$^{1,2}$ \quad Fernando Perez-Cruz $^{1,2}$\\ \\
$^1$Computer Science Department, ETH Zurich,  Zurich, Switzerland \\ $^2$Swiss Data Science Center, Zurich, Switzerland\\ \\
\texttt{nschneide@student.ethz.ch}\\
\texttt{shirin.goshtasbpour@inf.ethz.ch} \\
\texttt{fernando.perezcruz@sdsc.ethz.ch}
}

\usepackage{makecell}

\renewcommand{\vec}[1]{\boldsymbol{#1}}
\newcommand{\mat}[1]{\boldsymbol{#1}}

\begin{document}

\maketitle

\begin{abstract}
We propose a novel algorithm for data augmentation in nonlinear over-parametrized regression. Our data augmentation algorithm borrows from the literature on causality and extends the recently proposed Anchor regression (AR) method for data augmentation, which is in contrast to the current state-of-the-art domain-agnostic solutions that rely on the Mixup literature. Our Anchor Data Augmentation (ADA) uses several replicas of the modified samples in AR to provide more training examples, leading to more robust regression predictions. We apply ADA to linear and nonlinear regression problems using neural networks. ADA is competitive with state-of-the-art C-Mixup solutions. \footnote{Our Python implementation of ADA is available at: \url{https://github.com/noraschneider/anchordataaugmentation/}}
%In this paper, we propose an algorithm for data augmentation in nonlinear regression. Our data augmentation algorithm borrows from the literature on causality to extend the recently proposed Anchor regression (AR) method for data augmentation. Our Anchor Data Augmentation (ADA) uses several replicas of the modified samples in AR to provide more training examples that will result in more robust regression predictions. We apply ADA to linear and nonlinear regression problems using neural networks. We solve the regression problem using ordinary least squares closed-form solution, as well as stochastic gradient descent. The results of ADA are particularly strong when they solve over-parameterized systems, as in the classification counterpart.
\end{abstract}

\section{Introduction}
\label{sec:intro}

Data augmentation is one of the key ingredients of any successful application of a machine learning classifier. The first example that typically comes to mind is the in-depth description of the data augmentation in the now-famous Alexnet paper \cite{krizhevsky2012}. Data augmentation algorithms come in different flavors, and they mostly rely on the expectation that small perturbations, invariances, or symmetries applied to the input will not change the class label. That way, we can present ‘fresh new’ samples as alterations of the available examples for training. These transformations modify the input distribution to make the algorithm more robust for cases where the distribution of the test set may differ from that of the training set. We refer the reader to the related work section (\Cref{sec:relatedAD}) for an overview and description of different data augmentation strategies.

The literature for data augmentation in regression is slim. The paper on Mixup augmentation \cite{zhang2017mixup} proposes a simple and general scheme for data augmentation using convex combinations of samples. The authors only apply their data augmentation proposal to classification problems. They conjecture in the discussion that the application to regression is {\em straightforward}, however, this is not the case in practice. Mixup is theoretically analyzed in \cite{carratino2020mixup, zhang2020does} as a regularization technique for classification and regression problems. However, it is only illustrated in classification problems.

The Mixup algorithm has been extended to regression problems in \cite{hwang2021regmix, yao2022cmix}, in which the authors explain that Mixup cannot be blindly applied to regression problems. To our knowledge, these are the only two papers in which data augmentation for regression is proposed. RegMix \cite{hwang2021regmix} relies on a hard-to-train prior neural network controller before augmenting the data using a Mixup strategy. \textit{C-Mixup} \cite{yao2022cmix}, a method proposed more recently, solves some of the issues limiting the standard Mixup algorithm for regression problems. The authors propose to mix only closeby samples in the output space (i.e., samples which have close enough labels). This strategy is only valid when the target variables are monotonic with the input and is applied in a transformed space. The authors present comprehensive results in data augmentation for in-distribution generalization, task generalization and out-of-distribution robustness.

In this paper, we rely on the causality literature to provide a different avenue for augmenting data in regression problems. Causal discovery finds the causes of a response variable among a given set of observations or helps to recognize the causal relations between a set of variables \cite{Peters_book}. These causes allow us to understand how these relations will change if we were to intervene in a subset of the (input) variables or what would be the effect on the output. So, in general, the regression model will be robust to perturbations in the input variables making the prediction less sensitive to changes in the distribution of the test set. For example, the authors in \cite{Peters2016} use the invariance property for prediction to perform causal inference. 
In turn, Anchor Regression (AR) builds upon the causality literature to obtain robust regression solutions when the input variables have been perturbed \cite{rothenhausler2021anchor}. The procedure relies on anchor variables capturing the heterogeneity within a dataset and a parameter $\gamma$ that measures the deviation with respect to the least square solution. Once the values of the anchors are known, AR modifies the data and obtains the least square solution, as detailed in Section \ref{sec:anchorreg}.

In this paper, we propose {\bf Anchor Data Augmentation (ADA)} to augment the training dataset with several replicas of the available data. 
We use a simple clustering of the data to encode a homogeneous group of observations and use different values of $\gamma$ to robustify the solution to different strengths of potential distribution shifts. In every minibatch, we sample $\gamma$ from a predetermined range around $\gamma=1$. As AR was developed for linear regression, the data augmentation strategy needs to be modified for nonlinear regression accordingly. %It is important to emphasize: {\em ADA can be applied blindly to any regression problem}.
We validate ADA for in-distribution generalization and out-of-distribution robustness under the same conditions proposed in C-Mixup \cite{yao2022cmix}, as well as some illustrative linear and nonlinear regression examples. In the replicated experiments, ADA is competitive or superior to other augmentation strategies such as C-Mixup, although on some datasets the performance gain is marginal.

The rest of the paper is organized as follows: First, we provide background information in \Cref{sec:related}. We give a brief overview of related work on data augmentation in \Cref{sec:relatedAD} and summarize the key concepts on Anchor Regression in \Cref{sec:anchorreg}. Second, \Cref{sec:anchoraug} shows how we extend Anchor Regression and introduces ADA. \Cref{sec:experiments} reports empirical evidence that our approach can improve predictions, especially in over-parameterized settings. We conclude the paper in \Cref{sec:conc}.

\section{Background}
\label{sec:related}

\subsection{Data Augmentation}\label{sec:relatedAD}

Many different data augmentation methods have been proposed in recent years with several applications in mind. Still most augmentations we mention here use human-designed transformations based on domain knowledge which leave the target variable invariant. For instance, Cutout \cite{devries2017improved} is an image-specific augmentation technique that is successfully used to train models on CIFAR10 and CIFAR100 \cite{krizhevsky2009learning}, but was determined to be unsuitable for larger image datasets like ImageNet with higher resolution \cite{deng2009imagenet}. Other augmentation methods for images such as random crop, horizontal or vertical mirroring, random rotation, or translation \cite{lecun1998gradient,simonyan2014very} may similarly apply to a certain group of image datasets while being inapplicable to others, e.g. datasets of digits and letters. 

In an attempt to automate the augmentation process and reduce human involvement, policy or search-based automated augmentation methods were developed. In AutoAugment \cite{cubuk2019autoaugment} a neural network is trained with Reinforcement Learning (RL) to combine an assortment of transformations in varying strengths to apply on samples of a given dataset and improve the model accuracy. Methods such as RandAugment \cite{cubuk2020randaugment}, Fast AutoAugment \cite{lim2019fast}, UniformAugment \cite{lingchen2020uniformaugment} and TrivialAugment \cite{muller2021trivialaugment} aim at reducing the cost of the pretraining search phase in automated augmentation with randomized transformations and reduced search space. %, continuous action space and differentiable transformations (Faster AutoAugment \cite{hataya2020faster} and DADA \cite{li2020differentiable}).

Alternatively, in order to adapt the augmentation policy to the model during training, Population-Based Augmentation \cite{ho2019population} and Online Hyperparameter Learning \cite{lin2019online} use multiple data augmentation workers that are updated using evolutionary strategies and RL, respectively. 
Adversarial AutoAugment \cite{zhang2019adversarial} and AugMax \cite{wang2021augmax} optimize for the augmentation policy that deteriorates the training accuracy and improves its robustness. DivAug \cite{liu2021divaug} finds the policy which maximizes the diversity of the augmented data.

Having a separate search phase for optimal augmentation policy is computationally expensive and may exceed the required computation to train the downstream model \cite{xu2022universal, cubuk2020randaugment}. In addition, these methods and their online counterparts need to be trained separately on every single dataset. While OnlineAugment \cite{tang2020onlineaugment} and DDAS exploit meta-learning to avoid this problem, they still rely on a set of predefined class invariant transformations that require domain-specific information.

Generic transformations such as Gaussian or adversarial noise \cite{devries2017improved, taylor2018improving, lakshminarayanan2017simple} and dropout \cite{bouthillier2015dropout} are also effective in expanding the training dataset. Generative models such as Generative Adversarial Networks (GAN) \cite{goodfellow2020generative} and Variational Auto-Encoders (VAE) \cite{kingma2013auto} are trained in \cite{antoniou2017data,chadebec2021data, tang2020onlineaugment} to synthesize samples close to the low dimensional manifold of the data for classification. 

Mixup \cite{zhang2017mixup} is a popular data augmentation using a convex combination of pairs of samples from different classes and their softened labels for augmentation. Mixup is only evaluated on classification problems, even though it is claimed that the application to regression is straightforward. Various extensions of Mixup have been proposed to prevent data manifold intrusion \cite{verma2019manifold}, use more complex mixing strategies \cite{yun2019cutmix, liu2022automix} or account for saliency in augmented samples \cite{kim2021co, kim2020puzzle}. These methods were predominantly designed to excel in classification tasks. In particular, Mixup for regression was studied in \cite{carratino2020mixup, zhang2020does, hwang2021regmix, yao2022cmix} but it was reported to adversely impact the predictions in regression problems when misleading augmented samples are generated from a pair of faraway samples.

\subsection{Anchor Regression}
\label{sec:anchorreg}

We summarize the key concepts of Anchor Regression (AR) as presented in \cite{rothenhausler2021anchor}. Let $X\in\X$ and $y\in\Y$ be the predictors and target variables sampled from distribution $(X,y)\sim P_\text{train}$, $\X\subseteq \R^d$ and $\Y \subseteq \R$. Traditionally, a causal framework models the relation of $y$ and $X$ to accurately predict the value of $y$ under given interventions or arbitrary perturbations on $X$. A commonly held assumption is that the underlying causal relation among variables remains the same while the sampling distribution $P_\text{train}$ is altered by the intervention shift or the applied perturbation. For instance, if the distribution $P_\text{train}$ is induced by an unknown linear causal model, then the causally optimal parameters can be expressed as the solution to the optimization problem:
\begin{align}\label{eq:minimax}
    b_\text{causal} = \argmin_b \max_{P\in\mathcal P} \mathbb E_P[(y - X^Tb)^2],
\end{align}
where $\mathcal P$ is the class of distributions containing all interventions on components of $X$ \cite{rojas2018invariant}. Therefore, causal parameters provide distributionally robust predictions that are optimal under the intervention in $\mathcal P$. In comparison, Ordinary Least Squares (OLS): 
\begin{align}\label{eq:ols}
    b_\text{OLS} = \argmin_b \mathbb E_{P_\text{train}}[(y - X^Tb)^2],
\end{align}
may lead to arbitrarily large predictive errors on distributions in $\mathcal P$. On the other hand, on $P_\text{train}$, causal parameters $b_\text{causal}$ lead to conservative predictions, while $b_\text{OLS}$ presents optimal least squared performance.

To trade-off predictive accuracy on the training distribution with distribution robustness and to enforce stability over statistical parameters, AR \cite{rothenhausler2021anchor,buhlmann2020invariance} proposes to relax the regularization in the optimization problem in \eqref{eq:minimax} to a smaller class of distributions $\mathcal P$. %The authors suggest using the class of distributions where interventions are applied to a specific subset of $X$ or the most frequent or important directions of perturbations are considered within a certain strength range. 

Assume that $X$ and $y$ are centered and have finite variance. We use $A\in\R^q$ (called anchors) to denote the exogenous (random) variables in $X$ which generate heterogeneity in $y$. We further denote the $L_2$-projection on the linear span of the components of $A$ with $\text{P}_A$ and $\text{Id}(y) = y$. Under linear assumption between $A$ and $(X, y)$, we can write the relaxed optimization problem as:
\begin{align}\label{eq:anchor}
    b_{\gamma, A} = \argmin_b \mathbb E_{P_\text{train}}[((\text{Id} - \text{P}_A)(y - X^Tb))^2]+ \gamma\mathbb E_{P_\text{train}}[(\text{P}_A(y - X^Tb))^2],
\end{align}
where $\gamma>0$ is a hyperparameter. The first term of the AR objective in Equation~\ref{eq:anchor} is the loss after ``partialling out" the anchor variable, which refers to first linearly regressing out $A$ from $X$ and $y$ and subsequently using OLS on the residuals. The second term is the well-known estimation objective used in the Instrumental Variable setting \cite{didelez2010assumptions}. Therefore, for different values of $\gamma$ AR interpolates between the partialling out objective ($\gamma =0$) and the IV estimator ($\gamma\rightarrow\infty$) and coincides with OLS for $\gamma = 1$. \cv{The authors show that the solution of AR optimizes a worst-case risk under shift-interventions on anchors up to a given strength. This in turn increases the robustness of the predictions to distribution shifts at the cost of reducing the in-distribution generalization.}

In the finite-sample case with $n$ observations from $P_\text{train}$, let matrix $\mathbf X\in\R^{n\times d}$ contain the observations of $X$ and let $\mathbf Y\in\R^n$ be the vector of corresponding targets. Similarly, we denote the matrix containing the observations of $A$ with $\mathbf A\in\R^{n\times q}$ and we use $\mathbf\Pi_\mathbf A = \mathbf A\left(\mathbf A^T \mathbf A\right)^{\dagger} \mathbf A^T$ as the projection operator on the column space of the anchor matrix $\mathbf A$ where $\mathbf A^\dagger$ denotes the pseudo-inverse of matrix $\mathbf A$. Further, $\mathbf{I}$ denotes the identity matrix. Then, the finite-sample optimization regression problem can be written as
\begin{align}\label{eq:nanchor}
    \hat b_{\gamma, \mathbf A} = \argmin_b \| (\mathbf{I} - \mathbf \Pi_\mathbf A)(\mathbf Y - \mathbf Xb) \|_2^2+ \gamma\|\mathbf\Pi_\mathbf A(\mathbf Y - \mathbf Xb)\|_2^2.
\end{align}

The AR regression estimate $\hat b_{\gamma, \mathbf A}$  can be obtained by applying the OLS solution to a modified set of inputs and outputs:
\begin{align}
    \tilde {\mathbf X}_{\gamma, \mathbf A} &= \mathbf X + (\sqrt\gamma-1) \mathbf \Pi_\mathbf A \mathbf X \label{eq:augx}\\
    \tilde{\mathbf Y}_{\gamma, \mathbf A} &= \mathbf Y + (\sqrt\gamma-1) \mathbf \Pi_\mathbf A \mathbf Y\label{eq:augy}
\end{align}

%In the sequence, we use the bold capital letter notation $\mathbf C$ to denote a matrix and denote the projected matrix with $\mathbf C_\mathbf A = \mathbf \Pi_\mathbf A \mathbf C$ with subscript $\mathbf A$.

\section{Anchor Data Augmentation}
\label{sec:anchoraug}

In this section, we introduce Anchor Data Augmentation (ADA), a domain-independent data augmentation method inspired by AR. ADA does not require previous knowledge about the data invariances nor manually engineered transformations. As opposed to existing domain-agnostic data augmentation methods \cite{devries2017improved, taylor2018improving, verma2019manifold}, we do not require training of an expensive generative model, and the augmentation only adds marginally to the computation complexity of the training. %In contrast to C-Mixup \cite{yao2022cmix} we do not rely on monotonicity in the data and have a lower complexity of $O(qn)$, with $n$ being the sample size and $q$ the number of columns in $\mathbf{A}$. 
In addition, since ADA originates from a causal regression problem, it can be readily applied to regression problems. Even when ADA does not improve performance, its effect on performance remains minimal.

%First, we shortly explain the optimization procedure in the AR algorithm. To optimize the AR objective in finite-sample regime (\Cref{eq:nanchor}) for a fixed $\gamma$ and given anchor matrix, \cite{rothenhausler2021anchor} recommend the following two-stage least squares. In the first stage, we modify the training dataset with Equations (\ref{eq:augx}) and (\ref{eq:augy}) and in the second stage we compute the OLS parameters on the modified dataset $(\tilde{\mathbf X}_{\gamma, \mathbf A}, \tilde{\mathbf Y}_{\gamma, \mathbf A})$.

\cv{Data augmentation aims to introduce informative data in addition to the original dataset during the training procedure of the model to improve its generalization. Similar to AR, ADA employs a linear projection, given by the anchor variables $A$, to determine the most relevant perturbation directions based on the similarity of the samples. ADA  inherits the generalization properties from AR. In \cite{rothenhausler2021anchor}, the authors recommend that the anchor variable can be set as {\em an indicator of the datasets, where each dataset is a homogeneous set of observations}. A key insight of our work is that this can be achieved by clustering the data into $q$ clusters. The matrix $\mathbf{A}\in\mathbb{R}^{n\times q}$ is then constructed as an indicator matrix with a one-hot encoding of the assigned cluster index per row. For our experiments, we use \textit{k-means} clustering \cite{macqueen1967classification} to construct $\mathbf{A}$. Further, in AR, only one value for $\gamma$ is used, which should be chosen based on the desired strength of perturbations on test datasets, in comparison to the training dataset \citep{rothenhausler2021anchor}. We suggest that the value of $\gamma$ is sampled from a distribution with density $p(\gamma)$. In our experiments, we use a uniform distribution between $1/\alpha$ and  $\alpha$, where $\alpha>1$ is a hyperparameter to be tuned.}

\cv{ADA augments a sample $(\mathbf{X}^{(i)}, \mathbf{Y}^{(i)})$ by normalizing the original AR modifications (\ref{eq:augx} and \ref{eq:augy}) by $1 + (\sqrt\gamma-1)\sum_j(\mathbf\Pi_\mathbf A)^{(ij)}$ to unify the noise level across the augmentations independent of the value of $\gamma$, while approximately preserving the potentially nonlinear relation between $X$ and $y$ (see also \cref{sec:anchoraug_nonlinear}): 
\begin{align}
    \tilde{\mathbf{X}}^{(i)}_{\gamma, \mathbf A} &= \frac{\mathbf{X}^{(i)} + (\sqrt\gamma-1) (\mathbf \Pi_\mathbf A)^{(i)}\mathbf X}{1 + (\sqrt\gamma-1)\sum_j(\mathbf\Pi_\mathbf A)^{(ij)}}, \label{eq:modx}\\
    \tilde{\mathbf{Y}}^{(i)}_{\gamma, \mathbf A} &= \frac{\mathbf{Y}^{(i)} + (\sqrt\gamma-1) (\mathbf \Pi_\mathbf A)^{(i)}\mathbf Y}{1 + (\sqrt\gamma-1)\sum_j(\mathbf\Pi_\mathbf A)^{(ij)}},\label{eq:mody}
\end{align}
where we denote $(\mathbf M)^{(i)}$, $(\mathbf M)^{(:j)}$, and $(\mathbf M)^{(ij)}$ denote respectively the $i$-th row, the $j$-th column and $(i,j)$ component of some matrix $\mathbf M$. As is standard practice, we rely on stochastic gradient descent to optimize our (nonlinear) regressors and apply ADA on each minibatch rather than the entire dataset.}

\cv{
ADA combines samples from the same cluster and generates augmented samples along the data manifold. For a general $\mathbf A$, $\mathbf {\Pi_A}$ provides a “collective” mixing approach for the samples in a batch by determining a center, while $\gamma$ controls the extend of contraction or expansion of the augmented sample around this center. In particular, for a one-hot encoding matrix $\mathbf{A}$, $\mathbf{\Pi_A}^{(i)}\mathbf{X}$ defines the centroid of the cluster to which sample $i$ belongs. Then, the modified samples are located on the ray that originates from the centroid and goes through the original data point $(\mathbf{X}^{(i)}, \mathbf{Y}^{(i)})$. As $\gamma$ increases, the augmented samples move towards their corresponding centroid and specifically, for $\gamma = 1$ they coincide with the original samples. Furthermore, the cluster size, regulated by the number of clusters $q$, directly impacts the number of samples mixed together; with smaller clusters, fewer samples are combined. Applying ADA on each minibatch introduces further diversity and enhances robustness, because the composition of samples being mixed together and the value of $\gamma$ changes in each minibatch. In \Cref{app:anchoraug_hyper} we provide a detailed explanation and analysis of the impact of ADA hyperparameters, $q$ controlling the number of clusters and $\alpha$ controlling the range of values for $\gamma$. In \Cref{app:cmixupexp} we empirically show how regression performance varies with respect to these hyperparameters. }

\begin{SCfigure}
\centering
\includegraphics[width=0.5\textwidth]
{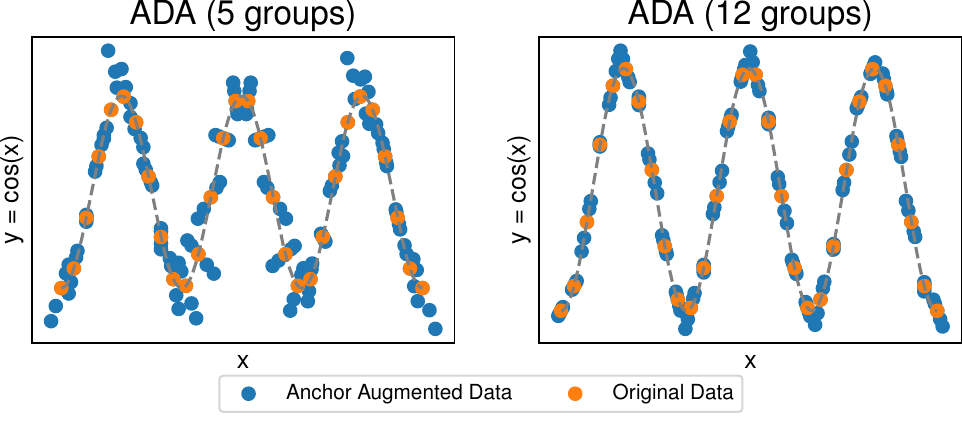}
\caption{Comparison of ADA augmentations on a nonlinear Cosine data model. For a larger partition size, ADA augmentations are more accurate due to the high local variability of the Cosine function. We used \textit{k-means} clustering to construct $\mathbf{A}$ and $\gamma \in \{1/2, 2/3, 1.0 3/2, 2.0\}$.}
\label{fig:cosine_numgroups}
\end{SCfigure}

\cv{In \Cref{fig:cosine_numgroups}, we visually illustrate the augmentation effects of ADA. We uniformly sampled 30 data points between $\pm3$ (i.e. $x_i\sim \mathcal{U}[-3,3]$) and set the corresponding target variable as $y_i= \cos(\pi x_i)$ without added noise. We then clustered this data in $q = 5$ and $q = 12$ groups using k-means and applied \cref{eq:modx} and \cref{eq:mody} to the 30 samples with $\gamma \in \{1/2, 2/3, 1, 3/2, 2\}$ resulting in 150 augmented data points.}

\subsection{Comparison to C-Mixup}
\cv{ADA can be interpreted as a generalized variant of C-Mixup \citep{yao2022cmix}. In C-Mixup samples are mixed in pairs, and the combination probability of each sample pair is given by the similarity of their labels, measured by a Gaussian kernel. Augmented samples are then obtained as the convex combination between the pair. In contrast, ADA allows mixing multiple samples based on their cluster membership and the resulting augmentations that may reside in the convex hull of the original samples of a cluster if $\gamma \geq 1$ or beyond it when $\gamma < 1$. In particular, ADA and C-Mixup augmentations would be similar if the anchor matrix $\mathbf A$ indicates pairs of samples weighted by the similarity of their labels and $\gamma > 1$.}

\subsection{Preserving nonlinear data structure}
\label{sec:anchoraug_nonlinear}
\cv{In the following we show that the scaled transformations in \cref{eq:augx}) and \cref{eq:augy} preserve the nonlinear relationship, so that we can use the modified pair $(\tilde{\mathbf X}_{\gamma,\mathbf A},\tilde{ \mathbf Y}_{\gamma, \mathbf A})$ to augment the dataset $(\mathbf X, \mathbf Y)$.} Let $(\mathbf{X^{(i)}}, \mathbf Y^{(i)})$ be the $i$th sample from $P_\text{train}$ corresponding to the $i$th row of $\mathbf X$ and $i$th component of $\mathbf Y$.
When the data has a nonlinear relation, 
\begin{align}\label{eq:nonlinregres}
    \mathbf Y^{(i)} = f_b(\mathbf{X^{(i)}}) + \eps^{(i)}
\end{align}
given the zero mean noise variable $\eps^{(i)}$, we can alter the anchor loss accordingly \cite{buhlmann2020invariance},
\begin{align}
    b_{\text{NONLIN},\gamma, A}, f_{\gamma, A} = %\nonumber\\
     \argmin_{b,f} &\mathbb E_{P_\text{train}}[((\text{Id} - \text{P}_A)(y - f_b(X)))^2] %\nonumber\\
    + \gamma\mathbb E_{P_\text{train}}[(\text{P}_A(y - f_b(X)))^2],\nonumber
\end{align}
The AR modification Equations~\ref{eq:augx} and \ref{eq:augy} do not preserve the nonlinear relation between the target and predictors,
\begin{align}
    \tilde{\mathbf{Y}}^{(i)} \neq f_b(\tilde{\mathbf{X}}^{(i)}) + \tilde\eps^{(i)}\nonumber
\end{align}
with another zero mean variable $\tilde \eps^{(i)}$ operating as the observation noise in the augmented data. Therefore, we propose to further extend the original AR and perform the data augmentation with scaled transformations to get the modified sample $(\tilde{\mathbf{X}}^{(i)}_{\gamma, \mathbf A},\tilde{\mathbf{Y}}^{(i)}_{\gamma, \mathbf A})$ which approximately preserves the nonlinear relationship of sample $(\mathbf{X^{(i)}},\mathbf{Y}^{(i)})$ as shown below.

%Using notation $(\mathbf A)^{(i)}$, $(\mathbf A)^{(:j)}$ and $(\mathbf A)^{(ij)}$ for $i$th row, $j$th column and the component at $i$th row and $j$th column of any matrix $\mathbf A$, we modify the transformations as follows:
%\begin{align}
%    \tilde {X}^{(i)}_{\gamma, \mathbf A} &= \frac{\mathbf{X^{(i)}} + (\sqrt\gamma-1) (\mathbf \Pi_\mathbf A)^{(i)}\mathbf X}{1 + (\sqrt\gamma-1)\sum_j(\mathbf\Pi_\mathbf A)^{(ij)}}, \label{eq:modx}\\
%    \tilde {y}^{(i)}_{\gamma, \mathbf A} &= \frac{y^{(i)} + (\sqrt\gamma-1) (\mathbf \Pi_\mathbf A)^{(i)}\mathbf Y}{1 + (\sqrt\gamma-1)\sum_j(\mathbf\Pi_\mathbf A)^{(ij)}} .\label{eq:mody}
%\end{align}

We can rewrite $\tilde{\mathbf{Y}}^{(i)}_{\gamma, \mathbf A} $ in Equation (\ref{eq:mody}) as
\begin{align}
    \tilde{\mathbf{Y}}^{(i)}_{\gamma, \mathbf A} = &\frac{f_b(\mathbf{X^{(i)}}) + (\sqrt\gamma-1) (\mathbf \Pi_\mathbf A)^{(i)}\mathbf F_b(\mathbf X)}{1 + (\sqrt\gamma-1)\sum_j(\mathbf\Pi_\mathbf A)^{(ij)}}%\nonumber\\
    + \tilde {\eps}^{(i)}_{\gamma, \mathbf A}\nonumber
\end{align}
where $\tilde {\eps}^{(i)}_{\gamma, \mathbf A}$ is a zero mean noise variable and $\mathbf F_b(\mathbf X) = [f_b(\mathbf X^{(1)}), ..., f_b(\mathbf X^{(n)})]^T$. In \Cref{app:anchoraug_taylor}, for continuously differentiable function $f$, we can use the first order Taylor expansion of $\tilde{\mathbf{Y}}^{(i)}_{\gamma, \mathbf A}$ around $\tilde{\mathbf{X}}^{(i)}_{\gamma, \mathbf A} $ to show that
\begin{align}\label{eq:nonlinmod}
    \tilde{\mathbf{Y}}^{(i)}_{\gamma, \mathbf A} \approx &f_b(\tilde{\mathbf{X}}^{(i)}_{\gamma, \mathbf A}) + \tilde {\eps}^{(i)}_{\gamma, \mathbf A} 
\end{align}
which approximately has the same nonlinear relation as the original model for small $\|\mathbf{X^{(i)}} - \tilde{\mathbf{X}}_{\gamma, \mathbf A}^{(i)}\|_2$ or $\|\sum_j(\mathbf \Pi_\mathbf A)^{(ij)}(\mathbf X^{(j)} - \tilde{\mathbf{X}}^{(i)}_{\gamma, \mathbf A})\|_2$.

With the one-hot partitioning matrix, $\mathbf{A}$ (introduced in the previous section), the approximation of the true nonlinear model becomes accurate in partitions with small diameter (where we define partition diameter as the maximum distance of two samples $\mathbf{X}^{(i)}$ and $\mathbf{X}^{(j)}$ in the same cluster).

\subsection{Algorithm}
\label{sec:algo}
\begin{wrapfigure}{r}{0.5\textwidth}
    \begin{minipage}{0.5\textwidth}
        \begin{algorithm}[H]
           \caption{ADA: Minibatch generation}
           \label{alg:ada}
            \begin{algorithmic}[1]
               \STATE {\bfseries Input:} $L$ training data points $(\mathbf{X}, \vec{Y})$; \\ prior distribution for $\gamma$: $p(\gamma)$\\ $L\times q$ binary matrix $\mathbf{A}$ with a one per row indicating the clustering assignment for each sample.\\
               \STATE {\bfseries Output:}  $(\Tilde{\mathbf{X}}, \Tilde{\vec{Y}})$
               \STATE Sample $\gamma$ from $p_{(\gamma)}$
               \STATE Projection matrix: $\mat{\Pi_{A}} \leftarrow \mathbf{A} (\mathbf{A}^T \mathbf{A})^{\dagger} \mathbf{A}^T$
               \FOR{$i=0$ {\bfseries to} row of $\mathbf{X}$}
               \STATE $\tilde{\mathbf{X}}^{(i)}_{\gamma, \mathbf A} \leftarrow \frac{\mathbf{X^{(i)}} + (\sqrt\gamma-1) (\mathbf \Pi_\mathbf A)^{(i)}\mathbf X}{1 + (\sqrt\gamma-1)\sum_j(\mathbf\Pi_\mathbf A)^{(ij)}} $
                \STATE $\tilde{\mathbf{Y}}^{(i)}_{\gamma, \mathbf A} \leftarrow \frac{\mathbf{Y}^{(i)} + (\sqrt\gamma-1) (\mathbf \Pi_\mathbf A)^{(i)}\mathbf Y}{1 + (\sqrt\gamma-1)\sum_j(\mathbf\Pi_\mathbf A)^{(ij)}}$
               \ENDFOR
               \STATE {\bfseries return} $(\Tilde{\mathbf{X}}_{\gamma, \mathbf A}, \Tilde{\vec{Y}}_{\gamma, \mathbf A})$
            \end{algorithmic}
        \end{algorithm}
    \end{minipage}
\end{wrapfigure}

Finally, in this section, we present the ADA algorithm step by step (\Cref{alg:ada}) to generate minibatches of data that can be used to train neural networks (or any other nonlinear regressor) by any stochastic gradient descent method. As discussed previously, we propose to repeat the augmentation with different parameter combinations for each minibatch. 

Given a centered training dataset $(\mathbf{X}, \mathbf{Y})$, its clustering assignment $\mathbf{A}$, and prior function $p(\gamma)$, the ADA minibatch algorithms takes $L$ random samples from the training set and its corresponding rows in $\mathbf A$ and outputs an $L$-sample mini-bath $(\Tilde{\mathbf{X}}_{\gamma, \mathbf A}, \Tilde{\mathbf{Y}}_{\gamma, \mathbf A})$. 

In order to do so, we first choose $\gamma$ according to the provided criterion $p(\gamma)$ (line 3). The corresponding projection matrix $\mathbf\Pi_A$ is computed from $\mathbf{A}$ (line 4). Finally, in lines five to seven, the transformation is applied according to Equations \ref{eq:modx} and \ref{eq:mody}.

%When the criteria $p_(\gamma)$ and $f_{\mathbf{A}}$ involve randomness the pairs $(\gamma, \mathbf{A})$ and therefore also $(\Tilde{\mathbf{X}}_{\gamma, \mathbf A}, \Tilde{\mathbf{Y}}_{\gamma, \mathbf A})$ are expected to differ in each execution. Hence, one can simply execute the algorithm multiple times to consider different parameter combinations for the augmented dataset. For large datasets, it can become intractable to do ADA on the entire data at once. A common approach in deep learning is to do computations on mini-batches, meaning smaller subsets of the dataset, instead of the entire dataset in order to reduce computational costs. Similarly, we suggest applying ADA for each minibatch of the data.}

\section{Experiments}
\label{sec:experiments}

%We experimentally demonstrate the effectiveness of Anchor Data Augmentation on synthetic and real world problems. In \Cref{sec:linearsyntheticdata}, we study synthetic linear data  - 

We experimentally investigate and compare the performance of ADA. First, we use ADA in an in-distribution setting for a linear regression problem (\Cref{sec:linearsyntheticdata}), in which we show that even in this case, ADA provides improved performance in the low data regime. Second, in \Cref{sec:housingdata}, we apply ADA and C-Mixup to the California and Boston Housing datasets as we increase the number of training samples. In the last two subsections, we replicate the in-distribution generalization (\Cref{sec:inDistGen}) and the out-of-distribution Robustness (\Cref{sec:outDistRob}) from the C-Mixup paper \cite{yao2022cmix}. In \cite{yao2022cmix} the authors further assess a task generalization experiment. However, the corresponding code was not publicly provided, and a comparison could not be easily made. 

\subsection{Linear synthetic data}
\label{sec:linearsyntheticdata}
Using synthetic linear data, we investigate if ADA can improve model performance in an over-parameterized setting compared to C-Mixup, vanilla augmentation, or classical expected risk minimization (ERM). Additionally, we analyze the sensitivity of our approach to the choice of $\gamma$ and the number of augmentations.

\textbf{Data:} The generated data follows a standard linear structure
\begin{align}
    \mathbf Y^{(i)} = \left(\mathbf X^{(i)}\right)^T b + b_{0} + \epsilon^{(i)}
\end{align}
with $\mathbf X^{(i)}, b \in \mathds{R}^{19}$ and $\mathbf Y^{(i)}, b_0, \epsilon^{(i)} \in \mathds{R}$. The parameters are sampled randomly from a Gaussian distribution $N(0,1)$. We sample $20$ different training datasets and one validation set with $\epsilon \sim \mathcal{N}(0,0.1^2)$, $\mathbf X^{(i)} \sim \mathbf{\mathcal{N}}(\vec{0}, \mathbf{I}_{19})$. For each training set, we take subsets with an increasing number of samples to evaluate the methods on different levels of data availability. The subsets are hierarchically constructed (i.e., meaning a smaller set is always a subset of a larger one). The validation set has 100,000 samples.

\textbf{Models and Comparisons:} We investigate and compare the impact of ADA using two different models with varying complexity: a linear Ridge regression and a multilayer perceptron (MLP) with one hidden layer and $10$ units with ReLU activation. %The MLP is optimized using Adam optimization \cite{kingma2014adam}. 
Using an MLP with more hidden layers shows similar results (see \Cref{app:linear} for details).

The ERM models only use the original data. We perform vanilla data augmentation by adding Gaussian noise $\epsilon' \sim N(0, 0.1^2)$ to the output leaving the input unchanged. Next, we apply C-Mixup with a bandwidth of $1$ and set the $\alpha$ of the Beta-distribution to $2$. Finally, we apply ADA with varying the number of obtained augmentations $k = \{10, 100\}$ and varying range of values for $\gamma$. To be precise, we define $\alpha \in \{2, 4, 6, 8, 10\}$ and specify $\beta_i = 1 + \frac{\alpha - 1}{k/2} \cdot i$ (with $i \in \{1, ..., k/2\}$) and $\gamma \in \{ \frac{1}{\alpha}, \frac{1}{\beta_{k/2 - 1}}, ... ,\frac{1}{\beta_1}, 1, \beta_{1},..., \beta_{k/2 - 1}, \alpha \}$. $\mathbf{A}$ is constructed using k-means clustering with $q=8$.  

For the Ridge regression model, we increase the dataset by a factor of $10$  by sampling from the respective augmentation methods and subsequently compute the regression estimators. In contrast, for the MLP, we implement the augmentation methods on a minibatch level. Specifically, we incorporate vanilla augmentation by adding Gaussian noise to each batch, apply C-Mixup after sampling from the beta distribution in each batch, and finally, apply ADA after sampling from the defined gamma values in each batch. 

% Scikit version
%The ERM models only use the original data. We perform vanilla data augmentation by adding Gaussian noise $\epsilon' \sim N(0, 0.1^2)$ to the output leaving the input unchanged. We use vanilla augmentation to increase the dataset by a factor $10$. Next, we apply C-Mixup with a bandwidth of 1 and set $\alpha$ of the Beta-distribution to $2$. We sample $10$ values from the Beta-distribution and augment the data accordingly. Finally, we apply ADA. We vary the number of obtained augmentations $k = \{10, 100\}$ and the range of values for $\gamma$. To be precise, we define $\alpha \in \{1.5, 2, 5, 10\}$ and specify $\beta_i = 1 + \frac{\alpha - 1}{k/2} \cdot i$ (with $i \in \{1, ..., k/2\}$) and $\gamma \in \{ \frac{1}{\alpha}, \frac{1}{\beta_{k/2 - 1}}, ... ,\frac{1}{\beta_1}, 1, \beta_{1},..., \beta_{k/2 - 1}, \alpha \}$. $\mathbf{A}$ is constructed using k-means clustering with $q=2$.

\begin{figure}[ht]
    \centering
    \includegraphics[width=\textwidth]
    {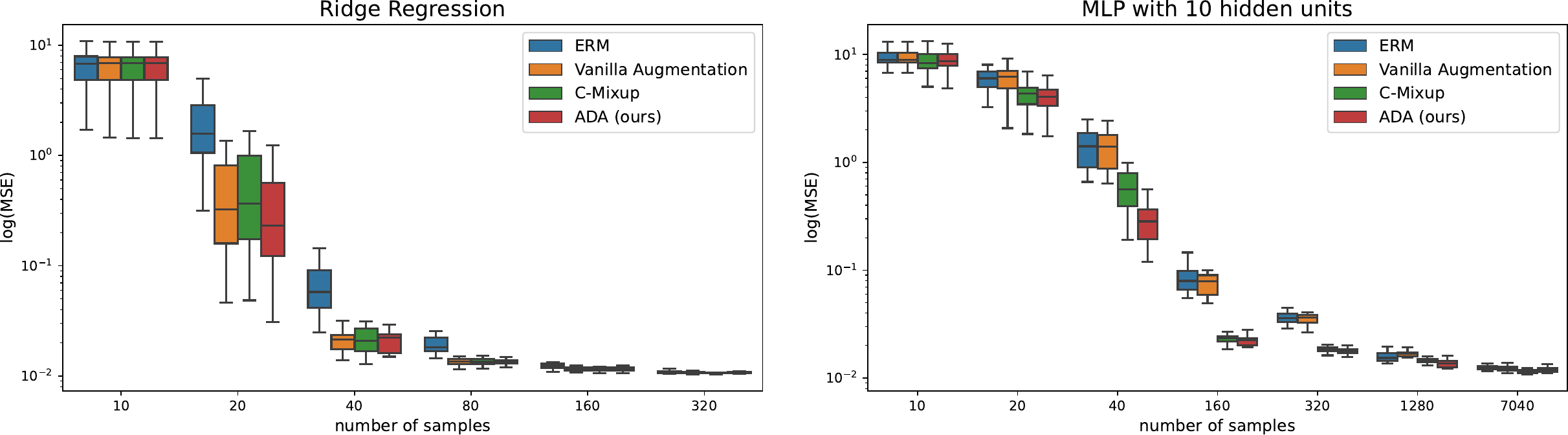}
    \caption{Mean Squared Error for Ridge Regression model and MLP model with varying number of training samples. For Ridge regression, vanilla augmentation and C-Mixup generate $k = 10$ augmented observations per observations. Similarly, Anchor Augmentation generates $k = 10$ augmented observations per observation with parameter $\alpha = 10$.}
    \label{fig:lin_result_baseanchorvanilla}
\end{figure}

\textbf{Results:}
We plot our results in \Cref{fig:lin_result_baseanchorvanilla}. First, as expected, Ridge regression outperforms the MLP model. Second, when there is little data availability, using ADA decreases the test error compared to ERM. The effect diminishes when the training dataset is sufficiently large, and all models converge to the noise limit of $0.1^2$. Third, vanilla augmentation achieves similar results as ADA and C-Mixup for Ridge regression but not quite for the MLP. This suggests that ADA (and C-Mixup) are more meaningful than randomly adding noise and especially well suited for highly parameterized models as the MLP has almost $20$ times more parameters than Ridge regression. In real-world applications, the value of $\epsilon$ is usually unknown, and choosing $\epsilon'$ for vanilla augmentation is not trivial, especially when the number of samples is small. Fourth, we conclude that generating more augmentations ($100$ instead of $10$) further improves prediction error in vanilla and anchor augmentation (\Cref{app:experiments} \Cref{fig:lin_result_100_10}) and the effectiveness of anchor augmentation is further increased as the range for $\gamma$ is wider (\Cref{app:experiments} \Cref{fig:lin_result_alpha}). Finally, C-Mixup and ADA perform similarly with ADA having a tendency to achieve a lower test error. 

In summary, even in the simplest of cases, in which we should not expect gains from ADA (or C-Mixup), these data augmentation strategies provide gains in performance when the number of training examples is not sufficient to achieve the error floor. 

\subsection{Housing nonlinear regression}
\label{sec:housingdata}
We extend the results from the previous section to the California and Boston housing data and compare ADA to C-Mixup \cite{yao2022cmix}.  We repeat the same experiments on three different regression datasets. Results are provided in \Cref{app:housing} and also show the superiority of ADA over C-Mixup for data augmentation in the implemented experimental setup. 

\textbf{Data: } We use the California housing dataset \cite{kelleypace1997spatial} and the Boston housing dataset \cite{harrison1978hedonichousing}. The training dataset contains up to $n = 406$ samples, and the remaining samples are for validation. We report the results as a function of the number of training points.
%We split the dataset into training, validation, and test sets with sizes $\left[1032, 9804, 9804 \right]$ for California and $\left[400, 53, 53 \right]$ for Boston. Similarly to the previous experiments, we increase the number of training points used to fit a model.
% consists of $n = 20640$ observations and $p = 8$ predictors.
% has $n = 406$ observations and $p = 13$ predictors.
% Next, we standardize the predictors. 

\textbf{Models and comparisons:} We fit a ridge regression model (baseline) and train a MLP with one hidden layer with a varying number of hidden units with sigmoid activation. The baseline models only use only the original data. We train the same models using C-Mixup with a Gaussian kernel and bandwidth of $1.75$. We compare the previous approaches to models fitted on ADA augmented data. We generate $20$ different augmentations per original observation using different values for $\gamma$ controlled via $\alpha = 4$ similar to what was described in \Cref{sec:linearsyntheticdata}. The Anchor matrix is constructed using k-means clustering with $q = 10$. 

\begin{figure}[ht]
    \centering
    \includegraphics[width=0.8\textwidth]{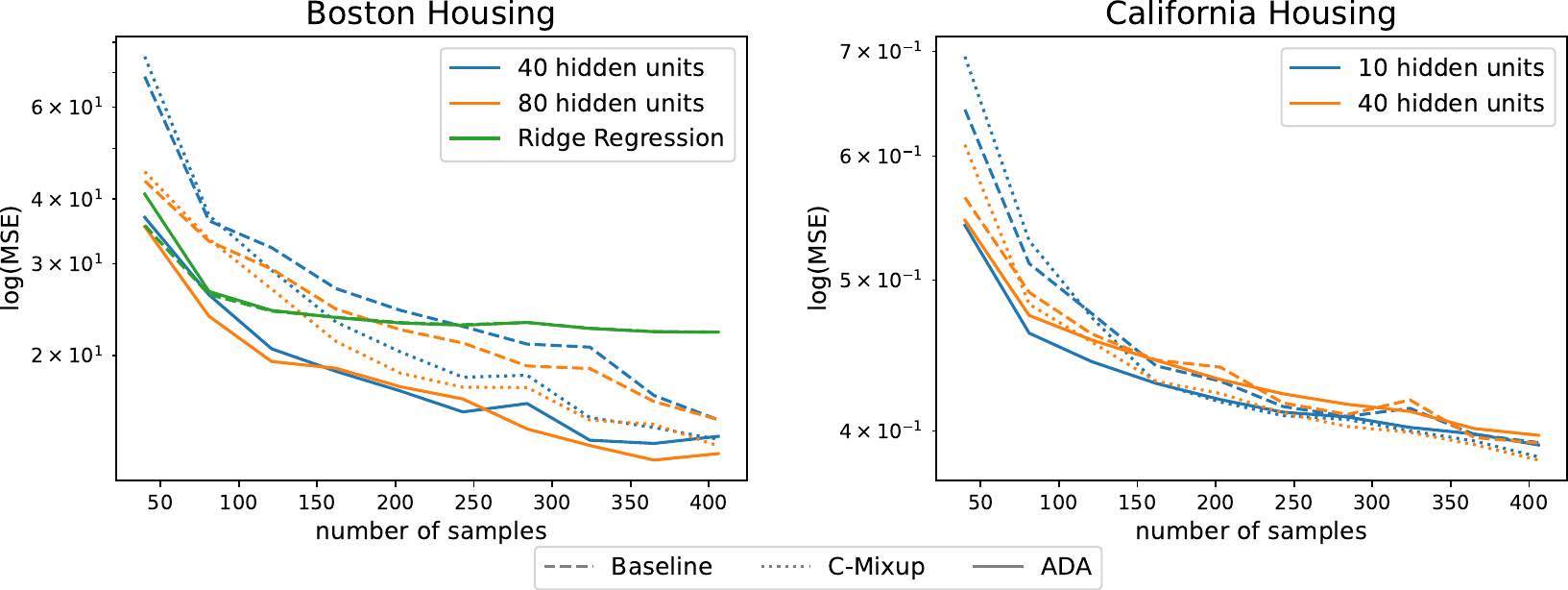}
    \caption{MSE for housing datasets averaged over $10$ different train-validation-test splits. On California housing Ridge regression performs much worse which is why it is not considered further (see \Cref{app:housing}).}
    \label{fig:housing_results}
    
\end{figure}

\textbf{Results:} \cv{We report the results in \Cref{fig:housing_results}}. First, we observe that the MLPs outperform Ridge regression suggesting a nonlinear data structure. Second, when the number of training samples is low, applying ADA improves the performance of all models compared to C-Mixup and the baseline. The performance gap decreases as the number of samples increases. When comparing C-Mixup and ADA, we see that using sufficiently many samples both methods achieve similar performance. While on the Boston data, the performance gap between the baseline and ADA persists, on California housing, the non-augmented model fit performs better than the augmented one when data availability increases. This suggests that there is a sweet spot where the addition of original data samples is required for better generalization, and augmented samples cannot contribute any further.

\subsection{In-distribution Generalization}
\label{sec:inDistGen}
In this section, we evaluate the performance of ADA and compare it to prior approaches on tasks involving in-distribution generalization. We use the same datasets as \cite{yao2022cmix} and closely follow their experimental setup. 

\textbf{Data:} We use four of the five in-distribution datasets used in \cite{yao2022cmix}. The validation and test data are expected to follow the same distribution as the training data. Airfoil Self-Noise (Airfoil) and NO2 \citep{kooperberg1997statlib} are both tabular datasets, whereas Exchange-Rate and Electricity \citep{lai2018modeling} are time series datasets. We divide the datasets into train-, validation- and test data randomly, as the authors of C-Mixup did. For Echocardiogram videos \citep{ouyang2020video} (the 5th dataset in \cite{yao2022cmix}), we could not replicate their preprocessing. 

\textbf{Models and comparisons: } We compare our approach, ADA, to C-Mixup \citep{yao2022cmix}, \cv{Local-Mixup \citep{baena2022preventing}}, Manifold-Mixup \citep{verma2019manifold}, Mixup \citep{zhang2017mixup} and classical expected risk minimization (ERM). Following the work of \cite{yao2022cmix}, we use the same model architectures: a three-layer fully connected network for the tabular datasets; and an LST-Attn \citep{lai2018modeling} for the time series.

We follow the setup of \citep{yao2022cmix} and apply C-Mixup, Manifold-Mixup, Mixup, and ERM with their reported hyperparameters and provided code. For the ADA and Local-Mixup experiments, we use hyperparameter tuning and grid search to find the optimal training (batch size, learning rate, and number of epochs), \cv{and Local-Mixup parameters (distance threshold $\epsilon$)} and ADA parameters (number of clusters, range of $\gamma$, and whether to use manifold augmentation). We provide a detailed description in Appendix \ref{app:cmixupexp}. The evaluation metrics are Root Mean Squared Error (RMSE) and Mean Averaged Percentage Error (MAPE). 

\textbf{Results:} We report the results in Table \ref{tab:realworlddata_indistr}. For full transparency, in the last row, we copy the results from \citep{yao2022cmix}. We can assess that ADA is competitive with C-Mixup and superior to the other data augmentation strategies. \cv{ADA consistently improves the regression fit compared to ERM.} Under the same conditions (split of data and Neural network structure), ADA is superior to C-Mixup. But, \cv{the degree of improvement is marginal on some datasets and} as we show in the last row, we could not fully replicate their results. The only data in which ADA is significantly better than C-Mixup and the other strategies is for the Airfoil data, in which ADA reduces the error by around 15\% with respect to the ERM solution. 

\vspace{-4mm}
\begin{table}[ht!]

    \caption{Results for in-distribution generalization. We report the average RMSE and MAPE of three different seeds. Standard deviations are reported in \Cref{app:cmixupexp}. The best results per column are printed in bold and the second-best results are underlined (not applicable to the last row).}
    \centering
    \begin{tabular}{lcccccccc}
    \toprule
         & \multicolumn{2}{c}{\textbf{Airfoil}} & \multicolumn{2}{c}{\textbf{NO2}} & \multicolumn{2}{c}{\textbf{Exchange- \newline Rate}} & \multicolumn{2}{c}{\textbf{Electricity}} \\
        \textbf{} & RMSE & MAPE & RMSE & MAPE & RMSE & MAPE & RMSE & MAPE \\ \midrule
        \textbf{ERM} & \underline{2.758} & 1.694 & 0.529 & 13.402 & 0.024 & 2.437 & 0.058 & 13.915 \\ 
        \textbf{Mixup} & 3.264 & 1.964 & 0.522 & 13.226 & 0.025 & 2.513 & \underline{0.058} & 13.839 \\ 
        \textbf{ManiMixup} & 3.092 & 1.871 & 0.528 & 13.358 & 0.025 & 2.541 & 0.058 & 14.031 \\ 
        \cv{\textbf{Local Mixup}} & 3.373 & 2.043 & 0.524 & 13.309 & \underline{0.021} & \underline{2.136} & 0.063 & 14.238\\
        \textbf{C-Mixup} & 2.800 & \underline{1.629} & \underline{0.516} & \textbf{13.069} & 0.024 & 2.431 & \textbf{0.057} & \underline{13.512} \\ \midrule
        \textbf{ADA} & \textbf{2.360} & \textbf{1.373} & \textbf{0.515} & \underline{13.128} & \textbf{0.021} & \textbf{2.116} & 0.059 & \textbf{13.464} \\ \midrule \midrule
            \textbf{C-Mixup} in \citep{yao2022cmix}  & 2.717 & 1.610 & 0.509 & 12.998 & 0.020 & 2.041 & 0.057 & 13.372 \\ 
        \bottomrule
    \label{tab:realworlddata_indistr}
    \end{tabular}
\vspace{-8mm}
\end{table}

\subsection{Out-of-distribution Robustness}
\label{sec:outDistRob}
In this section, we evaluate the performance of ADA and compare it to prior approaches on tasks involving out-of-distribution robustness. We use the same datasets as \cite{yao2022cmix} and closely follow their experimental setup. 

\textbf{Data:} We use four of the five out-of-distribution datasets used in \cite{yao2022cmix}. First, we use RCFashion-MNIST (RCF-MNIST) \citep{yao2022cmix}, which is a synthetic modification of Fashion-MNIST that models subpopulation shifts. Second, we investigate domain shifts using Communities and Crime (Crime) \citep{dua2019uci}, SkillCraft1 Master Table (SkillCraft) \citep{dua2019uci} \cv{ and Drug-target Interactions (DTI) \citep{huang2021therapeutics} all} of which are tabular datasets. For Crime, we use state identification, in SkillCraft we use "League Index", which corresponds to different levels of competitors, and in DTI we use year, as domain information. We split the datasets into train-, validation- and test data based on the domain information resulting in domain-distinct datasets. We provide a detailed description of datasets in Appendix \ref{app:cmixupexp}. \cv{Due to computational complexity, we could not establish a fair comparison on the satellite image regression dataset \citep{koh2021wilds} (the fifth dataset in \cite{yao2022cmix}), so we report some exploratory results in \Cref{app:cmixupexp}.}

\textbf{Models and comparisons:} As detailed in \Cref{sec:inDistGen}. Additionally, we use a ResNet-18 \citep{he2016deep} for RCF-MNIST and DeepDTA \citep{ozturk2018deepdta} for DTI, as proposed in \cite{yao2022cmix}.

\textbf{Results:} We report the RMSE and the "worst" domain RMSE, which corresponds to the worst within-domain RMSE for out-of-domain test sets in Table \ref{tab:realworlddata_ood}. Similar to \citep{yao2022cmix}, we report the $R$ value for the DTI dataset (higher values suggest a better fit of the regression model). For full transparency, in the last row, we copy the results from \citep{yao2022cmix}. We can assess that ADA is competitive with C-Mixup and the other data augmentation strategies. Under the same conditions (split of data and Neural network structure), ADA is superior to C-Mixup. But, \cv{the degree of improvement is marginal on some datasets and} as we show in the last row, we could not fully replicate their results. \cv{ADA is significantly better than C-Mixup and other strategies on the SkillCraft data, in which ADA reduces the error by around $15 \%$ compared to the ERM solution.} 
\vspace{-4mm}
 \begin{table}[ht!]
    \caption{Results for out-of-distribution generalisation. We report the average RMSE across domains in the test data and the "worst within-domain RMSE over three different seeds. For the DTI dataset, we report average R and "worst within-domain" R. Standard deviations are reported in \Cref{app:cmixupexp}. The best results per column are printed in bold and the second-best results are underlined (not applicable to the last row).}
    \centering
    \resizebox{0.9\textwidth}{!}{
   \begin{tabular}{lccccccc}
    \toprule
        & \textbf{RCF-MNIST} & \multicolumn{2}{c}{\textbf{Crimes}} & \multicolumn{2}{c}{\textbf{SkillCraft}} & \multicolumn{2}{c}{\textbf{DTI}} \\ 
        \textbf{} & \makecell{avg. \\ RMSE} & \makecell{avg. \\ RMSE} & \makecell{worst \\ RMSE} & \makecell{avg. \\ RMSE} & \makecell{worst \\ RMSE} & \makecell{avg. \\ R} & \makecell{worst \\ R} \\ \midrule
        \textbf{ERM} & 0.164 & 0.136 & 0.170 & 6.147 & \underline{7.906}  & \underline{0.483} & \underline{0.439} \\ 
        \textbf{Mixup} & 0.159 & 0.134 & 0.168 & 6.460 & 9.834 & 0.459 & 0.424 \\
        \textbf{ManiMixup} & \textbf{0.157} & \textbf{0.128} & \textbf{0.155} & \underline{5.908} & 9.264 & 0.474 & 0.431 \\
        \textbf{\cv{LocalMixup}} & 0.187 & 0.133 & 0.1590 & 7.251 & 10.996 & 0.470 & 0.433 \\
        \textbf{C-Mixup} & \underline{0.158} & 0.132 & 0.165 & 6.216 & 8.223 & 0.474 & 0.435 \\ \midrule
        \textbf{ADA} & 0.175 & \underline{0.130} & \underline{0.156} & \textbf{5.301} & \textbf{6.877} & \textbf{0.493} & \textbf{0.448} \\ \midrule\midrule
        \textbf{C-Mixup} in \citep{yao2022cmix} & 0.146 & 0.123 & 0.146 & 5.201 & 7.362 & 0.498 &  0.458 \\ 
\bottomrule
    \label{tab:realworlddata_ood}

    \end{tabular}
    }
\vspace{-10mm}
\end{table}

\section{Conclusion}
\label{sec:conc}

\cv{We introduced Anchor Data Augmentation (ADA), an extension of Anchor Regression for the purpose of data augmentation.  AR is a novel causal approach to increase the robustness in regression problems. In ADA, we systematically mix multiple samples based on a collective similarity criterion, which is determined via clustering. The augmented samples are modifications of the original samples that are moved towards or away from the cluster centroids based on the desired degree of robustness in AR. Our empirical evaluations across diverse synthetic and real-world regression problems consistently demonstrate the effectiveness of ADA, especially for limited data availability. ADA is competitive with or outperforms state-of-the-art data augmentation strategies for regression problems, even though the improvements are marginal on some datasets.}
%We proposed extending Anchor Regression for data augmentation. AR is a novel approach in the causality literature to do robust regression. Our ADA procedure shows that, by taking different replicas of the data, we can provide robust solutions. Our solutions are comparable or superior to state-of-the-art data augmentation strategies for regression problems.  

ADA can be applied to any regression setting, and we have not found any case in which the results were detrimental. To apply ADA, we only need to cluster our data and select a distribution for $\gamma$. We relied on vanilla {\em k-means}, and the results are robust with respect to the number of clusters. Other clustering algorithms might be more suitable for different applications. For setting the parameter $\gamma$, we used a uniform distribution. We believe a gamma distribution could be equally effective.

\section*{Broader Impact}

The purpose of data augmentation is to compensate for data scarcity in multiple domains where gathering and labeling data accurately by experts is impractical, expensive, or time-consuming. If applied properly, it can effectively expand the training dataset, reduce overfitting and improve the model's robustness, as was shown in the paper. However, It is important to note that the choice and combination of the data augmentation technique depends on the specific problem and using the wrong augmentation method may introduce additional bias to the model. 
%For instance, for natural images there is sufficient domain knowledge about the existing invariances and symmetries of the data to handcraft the corresponding transformations. The output of the transformations are predictable and the gains of the data augmentation methods are intuitively understandable. However, when more generic augmentation methods are used the effects are not immediately interpretable. To give an example, if we augment the data using synthetic samples from modern generative models we leave the model open to the imperfections of the generative model such as spurious or missing modes and manifold or density support mismatch. 
More generally, incorrect data augmentation can lead to the following problems: overfitting the augmented data, loss of important information, introduction of unrealistic patterns and imbalanced presentation of the data. Detecting emerging problems due to data augmentation may not be straightforward. In particular, the performance on a test distribution that matches the training data distribution may be misleading and the model's predictions should be used with caution on new data that reflects the potential distribution shifts or variations encountered in real-world.

%In this regard, we propose ADA which relies on AR for its augmentation transform. AR is specifically designed to account for the important shifts that are likely to occur in the test data through the Anchor matrix and we believe that ADA offers multiple benefits and opens up several possibilities for further exploration. For instance, a theoretical analysis of the method may enhance its interpretability. Further studies can shed light on and exploit the connection between the Anchor matrix and the invariances in the data for more efficient augmentation and better generalization. In addition we believe that a similar method can be used in transfer learning to provide new data from the target domain. An investigation on suitability of this method on critical decision making application is yet to follow.

% \begin{ack}
% Use unnumbered first level headings for the acknowledgments. All acknowledgments
% More information about this disclosure can be found at: \url{https://neurips.cc/Conferences/2020/PaperInformation/FundingDisclosure}.

% Do {\bf not} include this section in the anonymized submission, only in the final paper. You can use the \texttt{ack} environment provided in the style file to autmoatically hide this section in the anonymized submission.
% \end{ack}


\bibliography{references}
\bibliographystyle{plain}

%%%%%%%%%%%%%%%%%%%%%%%%%%%%%%%%%%%%%%%%%%%%%%%%%%%%%%%%%%%%%%%%%%%%%%%%%%%%%%%
%%%%%%%%%%%%%%%%%%%%%%%%%%%%%%%%%%%%%%%%%%%%%%%%%%%%%%%%%%%%%%%%%%%%%%%%%%%%%%%
% APPENDIX
%%%%%%%%%%%%%%%%%%%%%%%%%%%%%%%%%%%%%%%%%%%%%%%%%%%%%%%%%%%%%%%%%%%%%%%%%%%%%%%
%%%%%%%%%%%%%%%%%%%%%%%%%%%%%%%%%%%%%%%%%%%%%%%%%%%%%%%%%%%%%%%%%%%%%%%%%%%%%%%
\newpage
\appendix
\onecolumn
%\section{You \emph{can} have an appendix here.}
% comment for last run 
\section{Additional information for Anchor Data Augmentation}
\label{app:}
\subsection{Derivation of ADA for nonlinear data}\label{sec:app:nonlin}
\label{app:anchoraug_taylor}
In the following, we provide the more detailed derivation to \Cref{eq:nonlinmod}, which motivates the usage of the scaled transformation we use in ADA to obtain $(\tilde{\mathbf X}_{\gamma, \mathbf A}, \tilde{\mathbf Y}_{\gamma, \mathbf A})$. We use the same notation that was introduced in \Cref{sec:anchoraug}. As discussed in \Cref{sec:anchoraug}, we can write $\tilde{\mathbf{Y}}^{(i)}_{\gamma, \mathbf A} $ in Equation~\ref{eq:mody} as
\begin{align}
    % \tilde{\mathbf{Y}}^{(i)}_{\gamma, \mathbf A} = &\frac{f_b( \mathbf X^{(i)}) + (\sqrt\gamma-1) (\mathbf \Pi_\mathbf A)^{(i)}\mathbf F_b(\mathbf X)}{1 + (\sqrt\gamma-1)\sum_j(\mathbf\Pi_\mathbf A)^{(ij)}} + \tilde {\eps}^{(i)}_{\gamma, \mathbf A}\nonumber
\end{align}
for some noise variable $\tilde {\eps}^{(i)}_{\gamma, \mathbf A}$, where $\mathbf F_b(\mathbf X) = [f_b(\mathbf X^{(1)}), ..., f_b(\mathbf X^{(n)})]^T$. For differentiable function $f$ with continuous first-order derivative $\dot f$, we can use Taylor expansion around $\tilde{\mathbf{X}}^{(i)}_{\gamma, \mathbf A} $ of the nominator and get
\begin{align}
    f_b(X^{(i)}) + (\sqrt\gamma-1) (\mathbf \Pi_\mathbf A)^{(i)} \mathbf F_b(\mathbf X) = &  f_b(\tilde{\mathbf{X}}^{(i)}_{\gamma, \mathbf A}) + (\mathbf X^{(i)} - \tilde{\mathbf{X}}^{(i)}_{\gamma, \mathbf A})^T\dot f_b(\tilde{\mathbf{X}}^{(i)}_{\gamma, \mathbf A}) \nonumber\\
    & + (\sqrt\gamma-1) \sum_j(\mathbf \Pi_\mathbf A)^{(ij)}f_b(\tilde{\mathbf{X}}^{(i)}_{\gamma, \mathbf A}) \nonumber \\
    & + (\sqrt\gamma-1) \sum_j(\mathbf \Pi_\mathbf A)^{(ij)}(\mathbf X^{(j)} - \tilde{\mathbf{X}}^{(i)}_{\gamma, \mathbf A})\dot f_b(\tilde{\mathbf{X}}^{(i)}_{\gamma, \mathbf A}) \nonumber \\
    &+ \mathcal O(\| \mathbf X^{(i)} - \tilde{\mathbf{X}}_{\gamma, \mathbf A}^{(i)}\|_2\|\sum_j(\mathbf \Pi_\mathbf A)^{(ij)}(\mathbf X^{(j)} - \tilde{\mathbf{X}}^{(i)}_{\gamma, \mathbf A})\|_2)\nonumber\\
    = & \left(1 + (\sqrt\gamma-1)\sum_j(\mathbf \Pi_\mathbf A)^{(ij)}\right) f_b(\tilde {\mathbf X}_{\gamma, \mathbf A})\nonumber\\
    %& + \left((X_i - \tilde X_i) + (\sqrt\gamma-1) \sum_j(\Pi_A)_{ij}(X_j - \tilde X_i)\right)w\dot f(\tilde X_iw) \nonumber\\
    &+ \mathcal O(\| \mathbf X^{(i)} - \tilde{\mathbf{X}}_{\gamma, \mathbf A}^{(i)}\|_2\|\sum_j(\mathbf \Pi_\mathbf A)^{(ij)}( \mathbf X^{(j)} - \tilde{\mathbf{X}}^{(i)}_{\gamma, \mathbf A})\|_2)\nonumber
\end{align}
where in the second equality we use the fact that coefficient of $\dot f_b(\tilde {\mathbf{X}}^{(i)}_{\gamma, \mathbf A}) $ (in the second and fourth term) is zero for any $f_b$ due the definition of $\tilde{\mathbf{X}}^{(i)}$ in Equation~\ref{eq:modx} and therefore,
\begin{align}
    \tilde{\mathbf{Y}}^{(i)}_{\gamma, \mathbf A} \approx &f_b(\tilde{\mathbf{X}}^{(i)}_{\gamma, \mathbf A}) + \tilde {\eps}^{(i)}_{\gamma, \mathbf A} \nonumber
\end{align}
which is approximately similar to the original nonlinear model for small $\|\mathbf X^{(i)} - \tilde{\mathbf{X}}_{\gamma, \mathbf A}^{(i)}\|_2$ or $\|\sum_j(\mathbf \Pi_\mathbf A)^{(ij)}(\mathbf{X^{(j)}} - \tilde{\mathbf{X}}^{(i)}_{\gamma, \mathbf A})\|_2$.

\subsection{Additional information on hyperparameters of ADA}
\label{app:anchoraug_hyper}
In this section, we illustrate in a simple 1D example (i.e. cosine data used in Figure \ref{fig:cosine_numgroups}) how changes in the hyperparameter values modify the data and affect the achieved estimation. \cv{Additionally, we show in \Cref{app:cmixupexp} how ADA performance on real-world data is impacted by changes in the hyperparameter values.}

Having a fixed pair of $(\gamma, \mathbf A)$ enforces the model to learn the optimal parameters for a particular trade-off between performance on $P_\text{train}$ and predefined interventional distributions \cite{rothenhausler2021anchor}. Instead of limiting the regularization to a fixed pair of $(\gamma, \mathbf A)$ that performs well on a previously known set of interventions, we propose to optimize the loss simultaneously over a set of $\gamma\in [0,\infty)$ and different anchor matrices. In particular, we optimize the parameters on a mixture of essentially similar distributions to $P_\text{train}$ simultaneously. To reduce the anchor regression's regularization effect, we propose using a combination of the following methods to exploit the data invariances and avoid conservative predictions.

\paragraph{Anchor Matrices and Locality: } Anchor variable $A$ is assumed to be the exogenous variable that generates heterogeneity in the target and has an approximately linear relation with $(X, y)$ (see AR loss in Equation~\ref{eq:anchor}). It is recommended to choose the variable relying on expert knowledge about the features that the target has a higher dependence on or is possibly misrepresented in the dataset so that we encourage the robustness of the trained model against this type of discrepancy. After deciding the features, one way to construct the anchor matrix $\mathbf A$ is to partition the dataset according to the similarity of the features, using for example binning or clustering algorithms. Then we can fill the rows of $\mathbf A$ with a one-hot encoding of the partition index that each sample belongs to. 

\begin{figure}[H]
\begin{center}
    \includegraphics[width=0.5\textwidth]{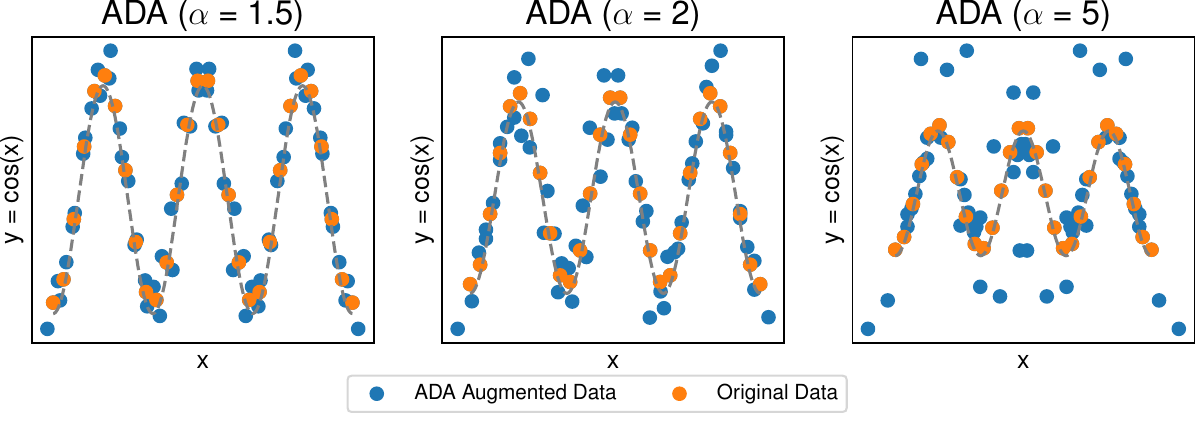}
\caption{ADA Augmented samples for varying ranges of $\gamma$ controlled via the parameter $\alpha$. We use k-means clustering into $q=5$ groups and augmented $2$ additional points per given point.}
\label{fig:cosine_rangegamma}
\end{center}
\end{figure}

\begin{figure}[!ht]
\begin{center}
    \includegraphics[width=0.5\textwidth]{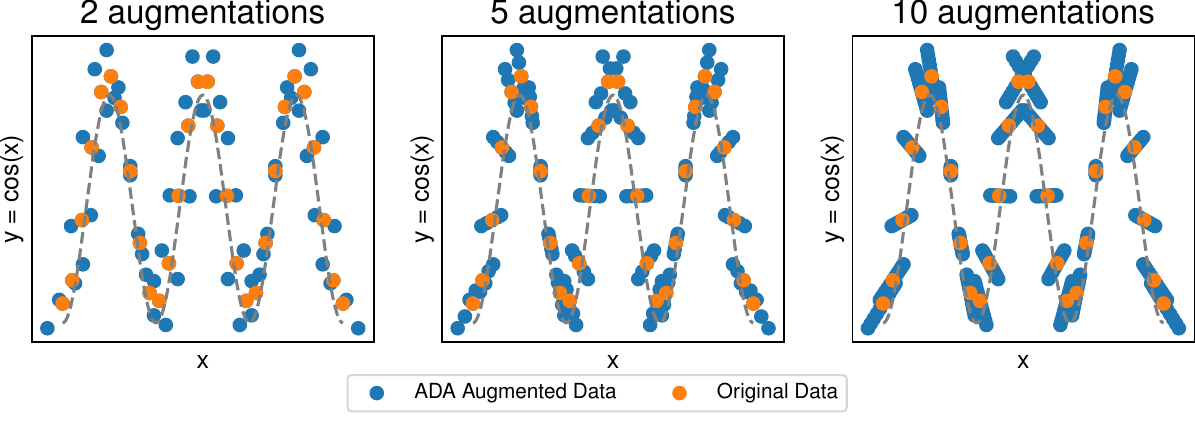}
\caption{ADA Augmented samples for varying numbers of parameter combinations. We use k-means clustering into $q=2$ groups $\alpha = 1.5$.}
\label{fig:cosine_numberaug}
\end{center}
\end{figure}

We use the following nonlinear Cosine data model as a running example to demonstrate more clearly how $\mathbf A$ is constructed and affects the augmentation procedure.
\begin{align}\label{eq:cos}
    \eps\sim\mathcal N(0, 0.1^2 \cdot \mathbf{I}),
    X\sim U(-3\pi, 3\pi),
    y = \cos(X^Tb) + \eps,
\end{align}
For illustration purposes, we use in Figures \ref{fig:cosine_rangegamma}, \ref{fig:cosine_numberaug} equidistant $x$ values as this reduces noise and emphasizes the effect of ADA parameters more.

Further, we note $a: \X\to\{1, ..., q\}$ that maps each sample $X\in\X$ to one of $q$ partitions and returns its index. % ($e_i\in\R^q$ denotes the unit vector which is one at the $i$th component and zero at the rest of $q-1$ components). 
For instance, with an equal width binning scheme one can partition the range of a feature map $g_k:\X\to[0,B]$ to $q$ parts and set $a(X)\coloneqq \argmin_{r\in\{1, ..., q\}} \{r: r/q \geq g_k(X)\}$. Using, equal size binning scheme, one would first sort $g_k(X^{(i)})$ for $i\in\{1, ..., n\}$ get the indices $o(g_k(X^{(i)}))$ accordingly and use $a(X^{(i)})\coloneqq \argmin_{r\in\{1, ..., q\}} \{r: rn/q \geq o(g_k(X))\}$. Similarly, it is possible to use a clustering algorithm such as k-means \cite{macqueen1967classification} to partition $\{X^{(i)}\}_i$ into hard clusters based on the similarity of each sample to cluster center $c_r\in\X$ for $r\in\{1, ..., q\}$ leading to $a(X) \coloneqq \argmin_{r\in\{1, ..., q\}} D(X, c_r)$ for some distance metric $D:\X\times \X\to [0,\infty)$.

With $\mathbf A$ constructed from the one-hot encoding of partition indices of samples, the $\mathbf \Pi_\mathbf A$ operator returns the average value of the projected values in the same group as each sample. 
\begin{align}
    (\mathbf \Pi_\mathbf A)^{(i)}\mathbf X &= \frac{1}{n_r}\sum_{j:a(\mathbf X^{(j)})=a(\mathbf X^{(i)})} \mathbf X^{(j)} \text{  and}\nonumber\\
    (\mathbf \Pi_\mathbf A)^{(i)} \mathbf Y &= \frac{1}{n_r}\sum_{j:a(\mathbf X^{(j)})=a(\mathbf X^{(i)})} \mathbf Y^{(j)},\nonumber
\end{align}
where $n_r$ is the size of group with index $r = a(\mathbf X^{(i)})$. Getting weighted averages of partition samples is also straightforward by scaling the one-hot encodings of group indices with the squared root of the desired weights. 

\begin{figure}[H]
\begin{center}
    \includegraphics[width=0.5\textwidth]{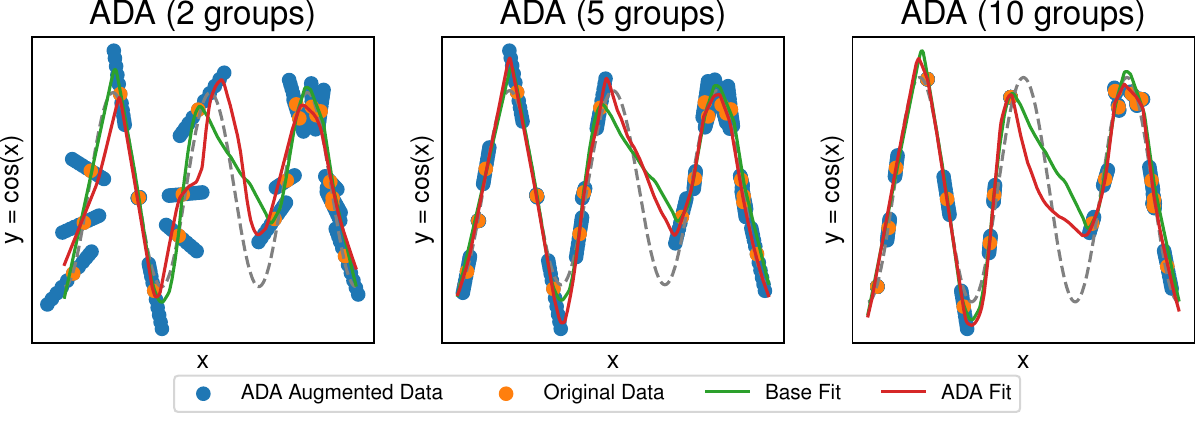}
\caption{Model predictions for models fit on the original data and ADA augmented data with varying partition sizes. On a hold-out validation set the base model has $MSE = 0.097$. The augmented model achieves MSEs of $0.124, 0.069, 0.079$, respectively. We use MLPs with architecture $\left[50, 50, 50, 50, 50\right]$ and ReLU activation function. The original data has $n=20$ points. We use k-means clustering, $\alpha = 2$, and augmented $10$ additional points per given point.}
\label{fig:cosine_numbergroups_fit}
\end{center}
\end{figure}

\paragraph{Partition Size and Number: } As was mentioned before, the target should have a high dependence on the anchor variable $A$. Specifically, with the partitioning scheme explained above, $\tilde{ \mathbf{X}}^{(i)}_{\gamma, \mathbf A}$ is constructed as a linear combination of $\mathbf X^{(i)}$ and the partition average with a target variable constructed in a similar manner. If the generative function $f$ varies significantly in a partition, the average value is going to flatten out the variations and decrease the heterogeneity of the augmented samples in that partition. For a smaller partition size, the augmented data is going to be close to the mean value $f$ and improve the optimization, however, partitions with a smaller number of samples will have a noisier estimation of the sample mean in each partition and deem the augmentation ineffective. We show the same effect of $q$ on the Cosine data model in Figure~\ref{fig:cosine_numbergroups_fit} for $\gamma$ set via $\alpha = 2$ (as described in \Cref{sec:linearsyntheticdata}) and a different number of groups when $g_k(X) = X$ and K-Means is used for partitioning the dataset. In the groups where $f$ is approximately linear, the augmentation line is approximately tangent to $f$, specifically when the clusters are small and the cluster average lies close to $\cos(X)$.

\begin{figure}[H]
\begin{center}
    \includegraphics[width=0.5\textwidth]{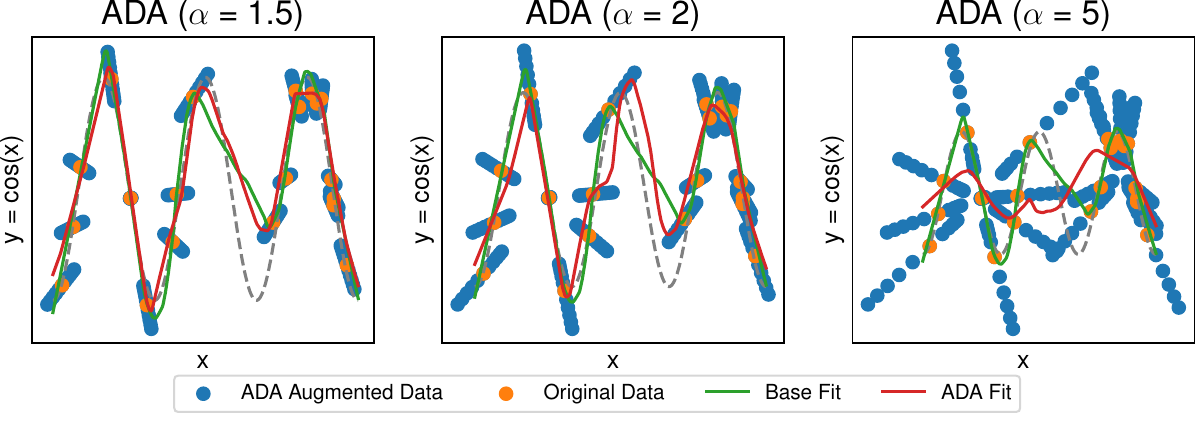}
\caption{Model predictions for models fit on the original data and ADA augmented data with different ranges of $\gamma$ controlled via the parameter $\alpha$. On a hold-out validation set the base model has $MSE = 0.097$. The augmented model fits achieve MSEs of $0.083, 0.124, 0.470$, respectively. We use MLPs with architecture $\left[50, 50, 50, 50, 50\right]$ and ReLU activation function. The original data has $n=20$ points. We use k-means clustering into $q=2$ groups and augmented $10$ additional points per given point.}
\label{fig:cosine_rangegamma_fit}
\end{center}
\end{figure}

\paragraph{Values of $\gamma$: } For $\gamma\in[0,\infty)$, the transformations in Equation~\ref{eq:modx} and \ref{eq:mody} defines a line passing through $(\mathbf X^{(i)}, \mathbf Y^{(i)})$ and the group average $((\mathbf \Pi_\mathbf A)^{(i)}\mathbf X, (\mathbf \Pi_\mathbf A)^{(i)}\mathbf Y)$. As $|\gamma - 1|$ grows the augmented sample gets further away from $\mathbf X^{(i)}$ and in large groups this may result in misleading augmentation. Therefore, when group diameter is large it is important to keep $\gamma$ close to one. In Figure~\ref{fig:cosine_rangegamma} we show how varying $\gamma$ changes the efficacy of the augmented samples for the Cosine data model with $q=2$ groups. To be precise, we vary the range of $\gamma$ by defining a parameter $\alpha \in \{1.5, 2, 5, 10\}$. We further specify $\beta_i = 1 + \frac{\alpha - 1}{k/2} \cdot i$ (with $i \in \{1, ..., k/2\}$) where $k$ is the number of augmentations and finally $\gamma \in \left\{ \frac{1}{\alpha}, \frac{1}{\beta_{k/2 - 1}}, ... ,\frac{1}{\beta_1}, 1, \beta_{1},..., \beta_{k/2 - 1}, \alpha \right\}$. Additionally, we provide a baseline and an augmented model fit in \Cref{fig:cosine_rangegamma_fit} with different values for $\gamma$.

\begin{figure}[!ht]
\begin{center}
    \includegraphics[width=0.5\textwidth]{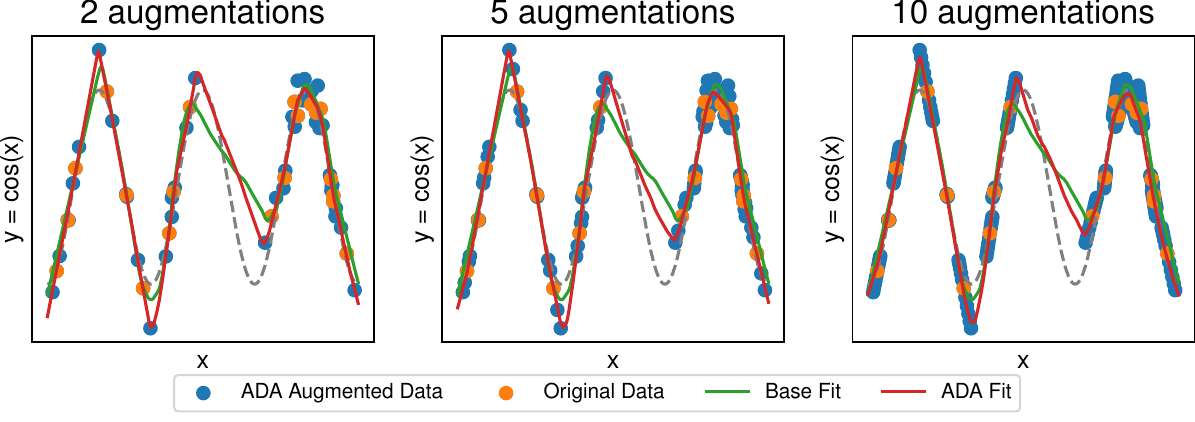}
\caption{Model predictions for models fit on the original data and ADA augmented data with a different number of parameter combinations (equal number of augmentations). On a hold-out validation set the base model has $MSE = 0.097$. The augmented model fits achieve MSEs of $0.470, 0.071, 0.057$, respectively. We use MLPs with architecture $\left[50, 50, 50, 50, 50\right]$ and ReLU activation function. The original data has $n=20$ points. We use k-means clustering into $q=5$ groups and $\alpha = 2$.}
\label{fig:cosine_numberaug_fit}
\end{center}
\end{figure}

\paragraph{Number of augmentations: } For each anchor matrix $\mathbf A$ and $\gamma$ we can add $n$ new samples to the dataset. The addition of more augmented samples may not be beneficial as the optimization may overfit the approximations in the augmented data model in Equation~\ref{eq:nonlinmod}. In the Cosine data model this is specifically problematic when $X$ is close to multiples of $\pi$ as depicted in Figure~\ref{fig:cosine_numberaug}. \cv{ Additionally, we provide a baseline and an augmented model fit in \Cref{fig:cosine_numberaug_fit} with different number of augmentations.}


\cv{
As it is standard practice to use stochastic gradient descent methods for optimizing a regressor, we suggest applying ADA on each minibatch instead of the entire dataset. This avoids choosing a fixed numbers of augmentations. Furthermore, it adds diversity to the "mixing" behavior of ADA, because the samples that are being mixed change in each iteration. }

\section{Experiments}
\label{app:experiments}
\subsection{Linear synthetic data}
\label{app:linear}
\begin{figure}[ht!]
\begin{center}
    \includegraphics[width=0.5
\textwidth]{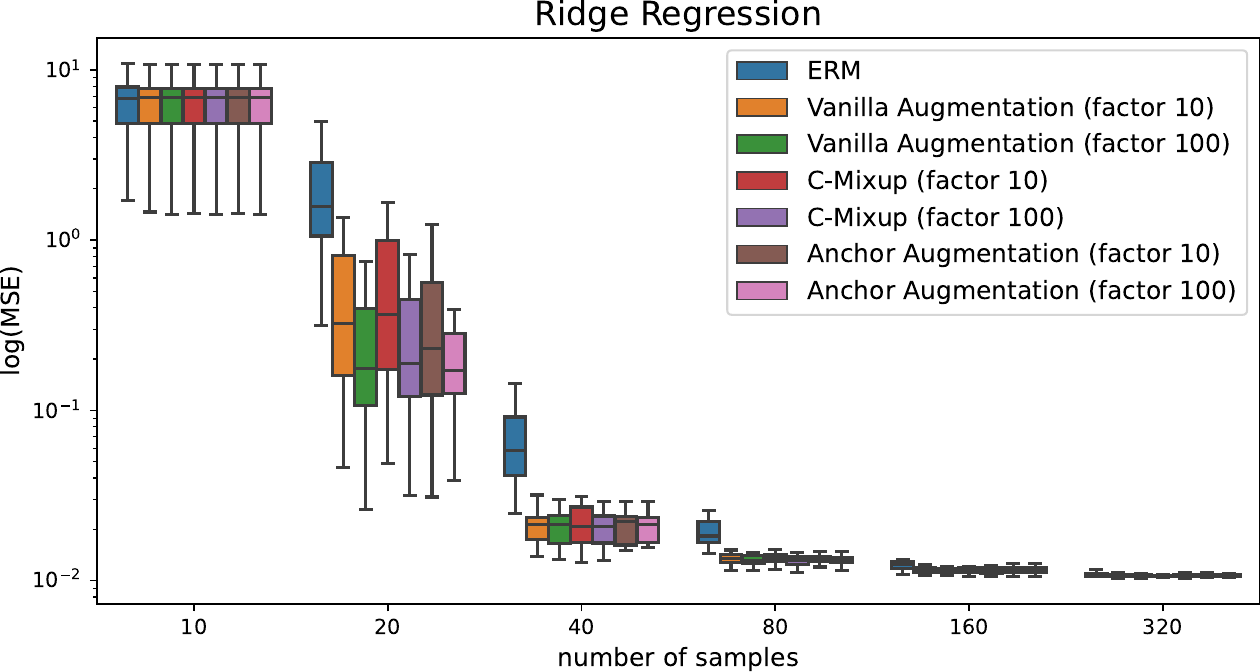}
\caption{Comparison of augmenting the synthetic linear dataset by a factor of $10$ and $100$. More augmentations achieve lower MSE on all methods. Here, anchor augmentation is performed for $\alpha = 8$.}
\label{fig:lin_result_100_10}
\end{center}
\end{figure}

In this section, we present more detailed results of the experiments on synthetic linear data (\Cref{sec:linearsyntheticdata}). First, \Cref{fig:lin_result_100_10} shows a comparison of using $10$ instead of $100$ additional augmentations per original sample using Ridge regression model. Performance increases when using $100$ instead of $10$ augmentations for all methods, as the resulting prediction error is lower.

\begin{figure}[ht!]
\begin{center}
    \includegraphics[width=0.85\textwidth]{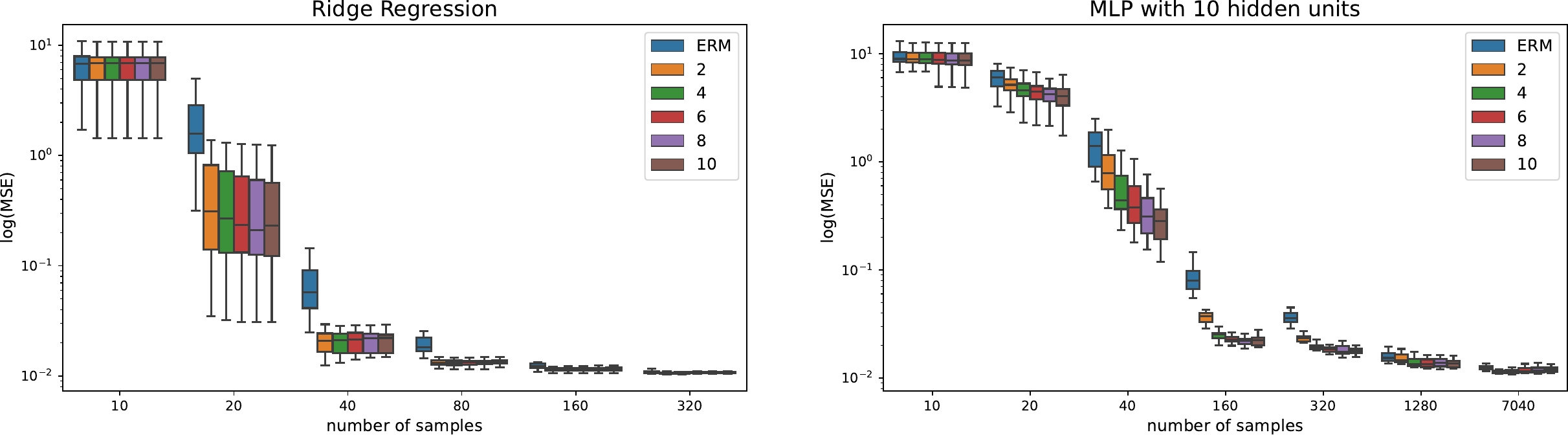}
\caption{Comparison of augmenting the synthetic linear dataset with different intervals of $\gamma$ controlled via $\alpha$. The ridge regression is not sensitive to the choice of $\alpha$, as different values result in a similar prediction error. Contrary, for the MLP a larger value of $\alpha$ is more effective.}
\label{fig:lin_result_alpha}
\end{center}
\end{figure}

Second, we report experimental results for using a wider interval for $\gamma$ values in \Cref{fig:lin_result_alpha}. The width is controlled via the parameter $\alpha$, as described in \Cref{sec:anchoraug}. While for ridge regression, the effectiveness of anchor augmentation is not sensitive to the choice of $\alpha$, the MLP model shows more sensitivity.

\begin{figure}[ht!]
\begin{center}
    \includegraphics[width=0.5\textwidth]{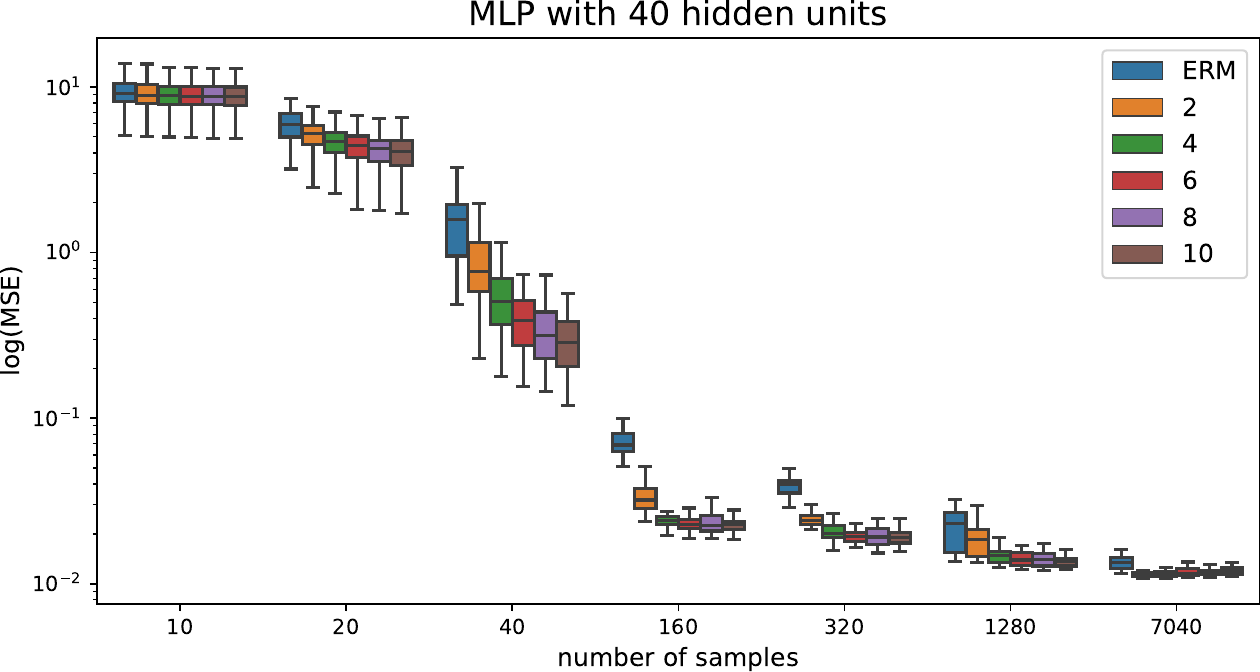}
\caption{Analysis of the sensitivity of ADA to the choice of $\alpha$ using an MLP with $40$ hidden units.}
\label{fig:lin_result_mlp40}
\end{center}
\end{figure}

Finally, we report results for using an MLP with $40$ hidden units in \Cref{fig:lin_result_mlp40}. The results are consistent with the results from the MLP with $10$ hidden unity.

\subsection{Additional results for real-world regression data}
\label{app:housing}
\begin{figure}[!ht]
\begin{center}
    \includegraphics[width=0.3\textwidth]{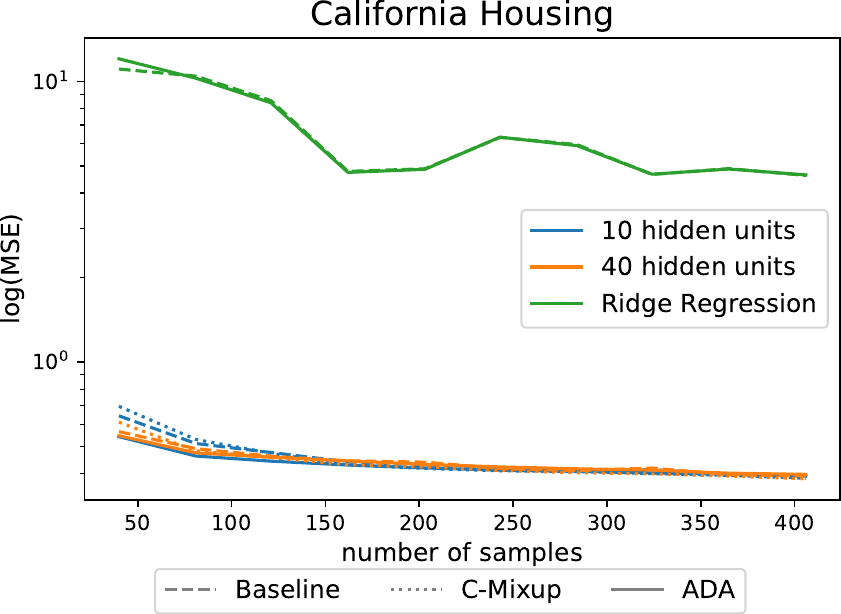}
\caption{MSE for California housing averaged over $10$ different train-validation-test splits.}
\label{fig:cal_full_result}
\end{center}
\end{figure}

In \Cref{fig:cal_full_result} we provide additional results showing, that the Ridge regression model performs worse on the California housing data. The experimental setting is the same as described in \Cref{sec:housingdata}. 

\subsection{Additional results on real-world data}
\label{app:extra_uci}

\begin{figure}[ht]
\begin{center}
    \includegraphics[width=0.75\textwidth]{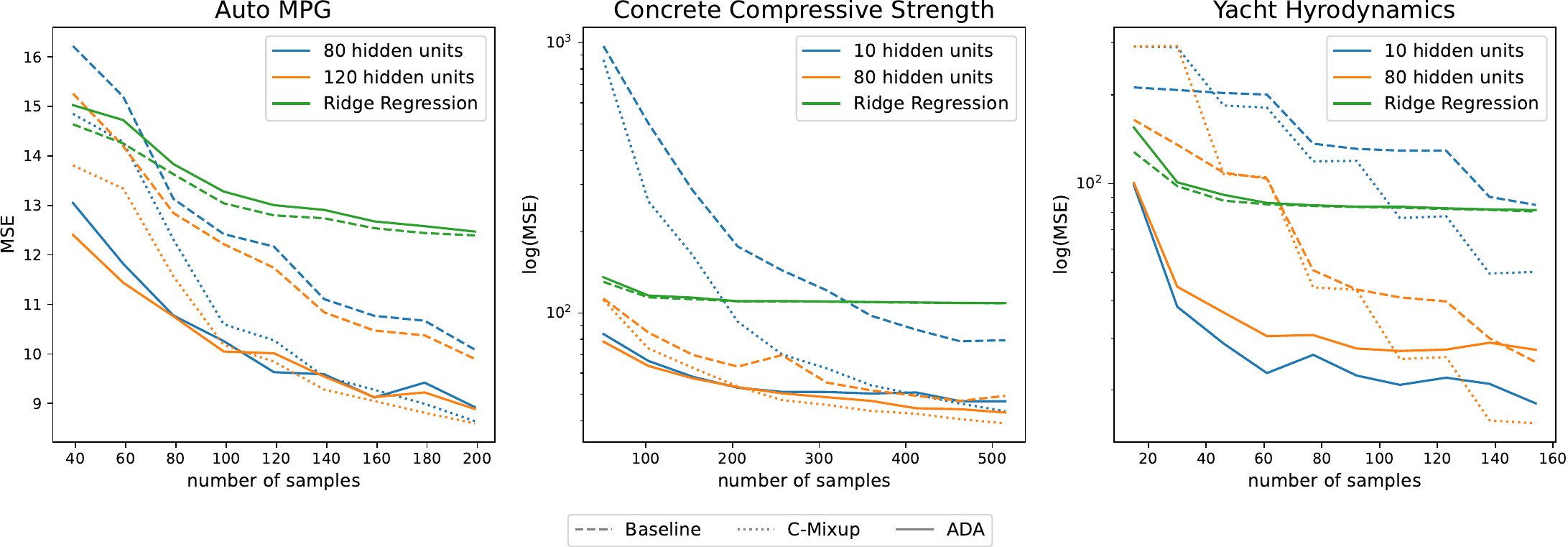}
\caption{MSE for housing datasets averaged over $9$ different train-validation-test splits.}
\label{fig:uci_result}
\end{center}
\end{figure}

In this section, we provide further experimental results on real-world regression problems. We use the following datasets from the UCI ML repository \cite{dua2019uci}: \textit{Auto MPG} (7 predictors), \textit{Concrete Compressive Strength}(8 predictors), and \textit{Yacht Hydrodynamics} (6 predictors). The experimental setting follows the one described in \Cref{sec:housingdata}, except that we use here to training, validation, and test datasets of relative sizes 50\%, 25\%, and 25\% respectively. We use MLPs with one layer and varying layer width and sigmoid activation. The models are trained using Adam optimization. We generate 9 different dataset splits and report the average prediction error in \Cref{fig:uci_result}. Similar to the results in \Cref{sec:housingdata}, ADA outperforms the baseline and C-Mixup especially when little data is available. The performance gap vanishes when more samples are available demonstrating the effectiveness of ADA in over-parameterized scenarios.

\subsection{Details: In-distribution Generalization and Out-of-distribution Robustness}
\label{app:cmixupexp}
In this section we present details for the experiments described in \cref{sec:inDistGen} and \Cref{sec:outDistRob}. We closely follow the experimental setup of \cite{yao2022cmix}. 

\subsubsection*{Data Description}
In the following, we provide a more detailed description of the datasets used for in-distribution generalization and out-of-distribution robustness experiments. 

\textbf{Airfoil \citep{dua2019uci}:} is a tabular dataset originating from aerodynamic and acoustic tests of two and three-dimensional airfoil blade sections. Each input has 5 features measuring frequency, angle of attack, chord length, free-stream velocity and suction side displacement thickness. The target variable is the scaled sound pressure level. As in \cite{yao2022cmix}, we additionally apply Min-Max normalization on the input featues and split the dataset into train (1003 samples), validation (300 samples) and test (200 samples) data. 

\textbf{NO2:} is a tabular dataset originating from a study where air pollution at a road is related to traffic volume and meteorological variables. Each input has 7 features measuring, the logarithm of the number of cars per hour, temperature $2$ meter above ground, wind speed, the temperature difference between $25$ and $2$ meters above ground, wind direction, hour of day and day number from 1st October 1 2001. The target variable is the logarithm of the concentration of NO2 particles, measured at Alnabru in Oslo, Norway. Following \cite{yao2022cmix}, we split the dataset into a train (200 samples), validation (200 samples) and test data (100 samples). 

\textbf{Exchange-Rate \cite{lai2018modeling}:} is a timeseries measuring the daily exchange rate of eight foreign countries including Australia, British, Canada, Switzerland, China, Japan, New Zealand and Singapore ranging from 1990 to 2016. The slide window size is 168 days, therefore the input has dimension $168 \times 8$ and the label has dimension $1 \times 8$. Following \cite{yao2022cmix, lai2018modeling} the dataset is split into training (4,373 samples), validation (1,518 samples) and test data (1,518 samples) in chronological order. 

\textbf{Electricity \cite{lai2018modeling}:} is a timeseries measuring the electricity consumption of 321 clients from 2012 to 2014. Similar to \cite{yao2022cmix, lai2018modeling} we converted the data to reflect hourly consumption. The slide window size is 168 hours, therefore the input has dimension $168 \times 321$ and the label has dimension $1 \times 321$. The dataset is split into training (15,591 samples), validation (5,261 samples) and test data (5,261 samples) in chronological order. 

\textbf{RCF-MNIST \cite{yao2022cmix}: } is rotated and colored version of F-MNIST simulating a subpopulation shift. The author rotate the images by a normalized angle $g\in \left[0, 1\right]$. In the training data they additionally color $80\%$ of images with RGB values $\left[g; 0; 1-g\right]$ and $20\%$ of images with RGB values $\left[1-g; 0; g\right]$. In the test data, they reverse the spurious correlations, so $80\%$ of images are colored with RGB values $\left[1-g; 0; g\right]$ and the remaining are colored with $\left[g; 0; 1-g\right]$.

\textbf{Crime \cite{dua2019uci}:} is a tabular dataset combining socio-economic data from the 1990 US Census, law enforcement data from the 1990 US LEMAS survey, and crime data from the 1995 FBI UCR. Each input has $122$ features that are supposed to have a plausible connection to crime, e.g. the median family income or per capita number of police officers. The target variable is the per capita violent crimes, representing the sum of violent crimes in the US including murder, rape, robbery, and assault. Following \cite{yao2022cmix}, we normalize all numerical features to the range $\left[0.0, 1.0\right]$ by equal-interval binning method and we impute the missing values using the mean value of the respective attribute. The state identifications are used as domain information. In total, there are 46 distinct domains and the data is split into disjoint domains. More precise, the training data has $1,390 samples$ with $31$ domains, the validation data has $231$ sampels with $6$ domains and the test data has $373$ samples with $9$ domains. 

\textbf{SkillCraft \cite{dua2019uci}:} is a tabular dataset originating from a study which uses video game from real-time strategy (RTS) games to explore the development of expertise. Each input has 17 features measuring player-related parameters, e.g. the age of the player and Hotkey usage variables. Following \cite{yao2022cmix}, we use the action latency in the game as a target variable. Missing values are imputed using the mean value of the respective attribute. "League Index", which corresponds to different levels of competitors, is used as domain information. In total there are 8 distinct domains and the data is split into disjoint domains. More precise, the training data has $1,878$ samples with $4$ domains, the validation data has $806$ samples with $1$ domain and the test data has $711$ samples with $3$ distinct domains. 

\cv{\textbf{DTI \cite{huang2021therapeutics}:} is a tabular dataset where the target is the binding activity score between a drug molecule and the corresponding target protein. The input consists of $32,300$ features which represent a one-hot encoding of drug and target protein information. Following \cite{yao2022cmix}, we use "Year" as domain information with $8$ distinct domains. There are $38,400$ training, $13,440$ validation and $11,008$ test samples. }

\subsubsection*{Methods and Hyperparameters}

For ERM, Mixup, ManiMixup and C-Mixup, we apply the same hyperparameters as reported in the original C-Mixup paper \cite{yao2022cmix}. According to the authors they are already finetuned via a cross-validation grid search. The details can be found in the corresponding original paper. We rerun their experiments with the provided repository (https://github.com/huaxiuyao/C-Mixup) over three different seeds $\in \{0, 1, 2\}$. Furthermore, we finetune ADA and training hyperparameters using a grid search. The detailed hyperparameters for in-distribution generalization and out-of-distribution robustness are reported in Table \ref{tab:adahyper}. We apply ADA using the same seeds. 

\begin{table}[!ht]
    \centering
    \caption{Hyperparameters for ADA}
    \resizebox{\textwidth}{!}{
    \label{tab:adahyper}
    \begin{tabular}{lcccccccc}
    \toprule
        & \textbf{Airfoil} & \textbf{NO2} & \textbf{Exchange} & \textbf{Electricity} & \textbf{RCF-MNIST} & \textbf{Crime}  & \textbf{SkillCraft} & \textbf{DTI} \\\midrule
        \textbf{Architecture} & FCN3 & FCN3 & LST-Attn &  LST-Att & ResNet-18 & FCN3 & FCN3 &  DeepDTA \\
        \textbf{Learningrate} & 0.01 & 5e-4 &  5e-4 & 5e-4 & 7e-5 & 1e-4 & 0.001 & 1e-4 \\ 
        \textbf{Optimizer} & Adam & Adam & Adam & Adam & Adam & Adam & Adam & Adam \\
        \textbf{Batchsize} & 16 & 32 & 64 & 64 & 128 & 48 & 48 & 32 \\
        \textbf{Maximum Epoch} & 200 & 150 & 100 & 100 & 40 & 250 & 100 & 20\\  \midrule
        \textbf{$\mathbf{\alpha}$ (determines $\mathbf{\gamma}$)} & 2 & 3.5 & 1.125 & 1.125 & 3 & 2.5 & 4 & 3 \\ 
        \textbf{Number Groups} & 8 & 4 & 40 & 40 & 25 & 2 & 16 & 24  \\
        \textbf{Manifold} & 1 & 0 & 0 & 0 & 1 & 1 & 0  & 1\\\bottomrule
    \end{tabular}
    }
\end{table}

\begin{figure}[ht]
    \centering
    \caption{Results for different $\alpha$ values (upper row) and $q$ values (lower row). Results are reported of three different seeds $\in \{0, 1, 2\}$. For Airfoil and Electricity we report RMSE and for Crimes we report "worst within-domain" RMSE.}
    \begin{subfigure}[b]{0.8\textwidth}
        \centering
        \includegraphics[width=\textwidth]{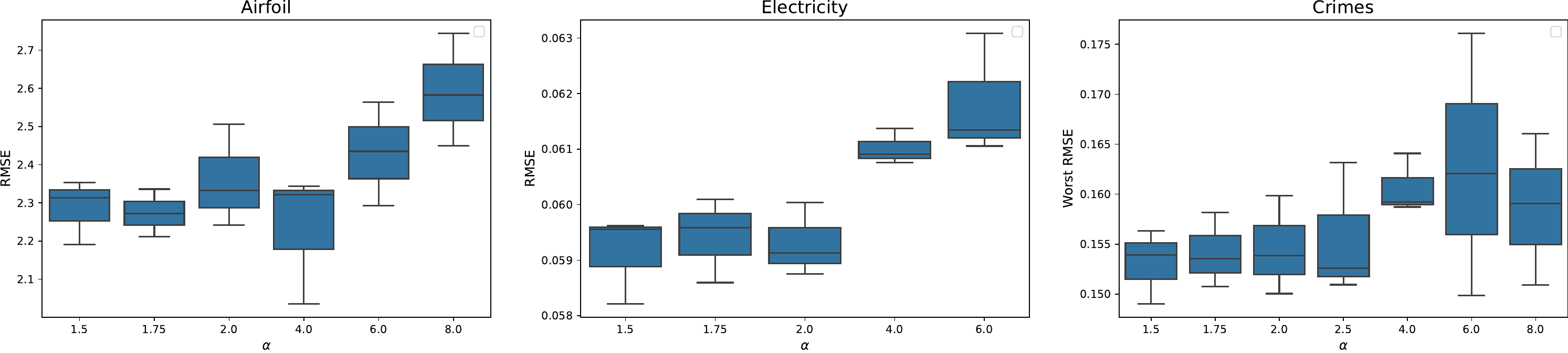}
    \end{subfigure}
    \hfill
    \begin{subfigure}[b]{0.8\textwidth}
        \centering
        \includegraphics[width=\textwidth]{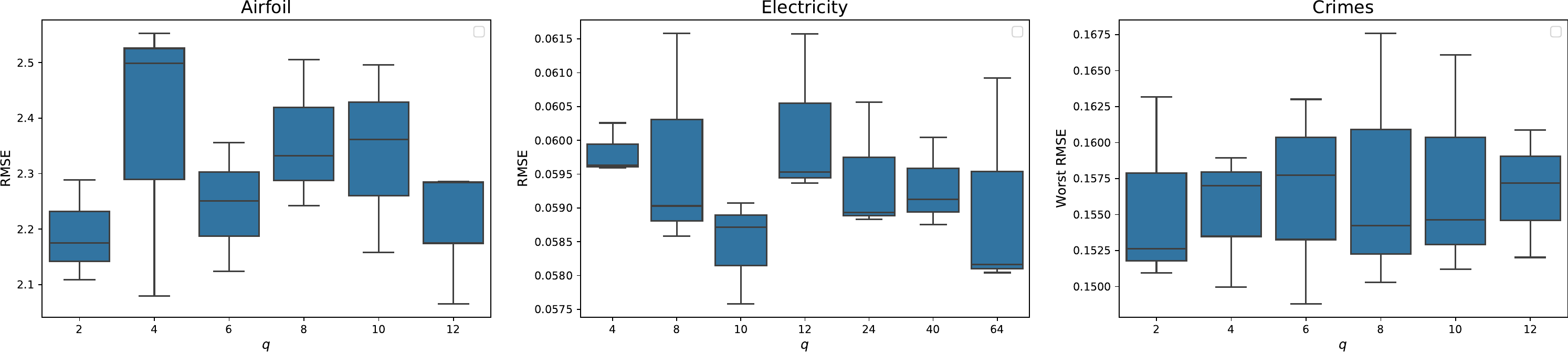}
    \end{subfigure}
    \label{fig:hyperparam_values}
\end{figure}

\cv{Furthermore, we provide the performance of ADA for different parameter parameter values to get a better understanding of their impact. We vary values for $q$, the number of clusters used in k-means clustering, and $\alpha$, the parameter that controls the range of $\gamma$-values on selected in-distribution and out-of-distribution datasets. Results are reported in \Cref{fig:hyperparam_values}.}

\subsubsection*{Detailed Results}
We report the results for in-distribution generalization experiments in \Cref{tab:detailedInDistr} and for out-of-distribution generalization experiments in \Cref{tab:detailedOutOfDistr}. \cv{Following \citep{yao2022cmix}, we further evaluated the performance of ADA and C-Mixup on the Poverty dataset \citep{koh2021wilds}, which contains  satellite images from African countries and the corresponding village-level real-valued asset wealth index. Again we closely followed the experimental setup, so for details we refer to \citep{yao2022cmix}. However, due to computational complexity, ADA hyperparameters are not tuned on this dataset. We use the same learningparameters as reported in \citep{yao2022cmix} and $q = 24$ and $\alpha = 2$. }

\begin{table}[!ht]
    \centering
    \caption{Detailed results for in-distribution generalization. We report the average RMSE and MAPE and the respective standard deviations over three seeds $\in \{0, 1, 2\}$.}
    \label{tab:detailedInDistr}
    \resizebox{0.9\textwidth}{!}{
    \begin{tabular}{lcccccccc}
    \toprule
        & \multicolumn{4}{c}{\textbf{Airfoil}} & \multicolumn{4}{c}{\textbf{NO2}} \\
        ~ & \multicolumn{2}{c}{RMSE} & \multicolumn{2}{c}{MAPE} & \multicolumn{2}{c}{RMSE} & \multicolumn{2}{c}{MAPE} \\
        ~ & mean & std & mean & std & mean & std & mean & std \\ \midrule
        \textbf{ERM} & 2.7582 & 0.1094 & 1.6942 & 0.0486 & 0.5294 & 0.0128 & 13.4019 & 0.3373 \\ 
        \textbf{Mixup} & 3.2637 & 0.1633 & 1.9645 & 0.1092 & 0.5220 & 0.0037 & 13.2260 & 0.1237 \\
        \textbf{ManiMixup} & 3.0922 & 0.1668 & 1.8712 & 0.0903 & 0.5277 & 0.0066 & 13.3579 & 0.2148 \\
        \textbf{Local Mixup} & 3.3727 & 0.1068 & 2.0426 & 0.0211 & 0.5242 & 0.0004 & 13.3087 & 0.1990 \\
        \textbf{C-Mixup} & 2.7997 & 0.2136 & 1.6289 & 0.1088 & 0.5157 & 0.0123 & 13.0688 & 0.3593 \\ \midrule
        \textbf{ADA} & 2.3601 & 0.1339 & 1.3730 & 0.0564 & 0.5147 & 0.0075 & 13.1277 & 0.1468 \\ \bottomrule
    \end{tabular}
    }
    \bigskip
    \centering
    \resizebox{0.9\textwidth}{!}{
    \begin{tabular}{lcccccccc}
    \toprule
        & \multicolumn{4}{c}{\textbf{Exchange-rate}} & \multicolumn{4}{c}{\textbf{Electricity}} \\
        ~ & \multicolumn{2}{c}{RMSE} & \multicolumn{2}{c}{MAPE} & \multicolumn{2}{c}{RMSE} & \multicolumn{2}{c}{MAPE} \\
        ~ & mean & std & mean & std & mean & std & mean & std \\ \midrule
        \textbf{ERM} & 0.0236 & 0.0065 & 2.4366 & 0.7142 & 0.0582 & 0.0002 & 13.9153 & 0.3410 \\
        \textbf{Mixup} & 0.0246 & 0.0058 & 2.5131 & 0.6667 & 0.0581 & 0.0002 & 13.8390 & 0.0539 \\
        \textbf{ManiMixup} & 0.0246 & 0.0065 & 2.5411 & 0.7417 & 0.0583 & 0.0004 & 14.0308 & 0.2174 \\
        \textbf{Local Mixup} & 0.0209 & 0.0046 & 2.1360 & 0.5851 & 0.0627 & 0.0054 & 14.2382 & 1.2349 \\
        \textbf{C-Mixup} & 0.0238 & 0.0061 & 2.4307 & 0.6819 & 0.0573 & 0.0003 & 13.5121 & 0.0979 \\ \midrule
        \textbf{ADA} & 0.0209 & 0.0060 & 2.1159 & 0.6889 & 0.0587 & 0.0008 & 13.4642 & 0.2956 \\
    \bottomrule
    \end{tabular}
    }
\end{table}

\begin{table}[!ht]
    \centering
    \caption{Detailed results for out-of-distribution generalization. We report the average and standard deviation of average and worst RMSE or R over three seeds $\in \{0, 1, 2\}$.}
    \label{tab:detailedOutOfDistr}
    \resizebox{\textwidth}{!}{
    \begin{tabular}{lcccccccccc}
    \toprule
        &  \multicolumn{2}{c}{\textbf{RCF-MNIST}} & \multicolumn{4}{c}{\textbf{Crime}} & \multicolumn{4}{c}{\textbf{SkillCraft}} \\
        \textbf{} & \multicolumn{2}{c}{avg. RMSE} & \multicolumn{2}{c}{avg. RMSE} & \multicolumn{2}{c}{worst RMSE} &\multicolumn{2}{c}{avg. RMSE} & \multicolumn{2}{c}{worst RMSE} \\
        \textbf{} & mean & std & mean & std & mean & std & mean & std & mean & std \\ \midrule
        \textbf{ERM} & 0.1636 & 0.0066 & 0.1356 & 0.0057 & 0.1698 & 0.0066 & 6.1473 & 0.4070 & 7.9064 & 0.3223 \\
        \textbf{Mixup} & 0.1585 & 0.0048 & 0.1341 & 0.0031 & 0.1681 & 0.0171 & 6.4605 & 0.4259 & 9.8338 & 0.9415 \\
        \textbf{ManiMixup} & 0.1572 & 0.0205 & 0.1283 & 0.0030 & 0.1554 & 0.0086 & 5.9080 & 0.3438 & 9.2643 & 1.0117 \\
        \textbf{Local Mixup} & 0.1873 & 0.0179 & 0.1325 & 0.0033 & 0.1590 & 0.0052 & 7.2514 & 0.4121 & 10.9957 & 0.5702 \\
        \textbf{C-Mixup} & 0.1579 & 0.0066 & 0.1320 & 0.0017 & 0.1647 & 0.0045 & 6.2156 & 0.3822 & 8.2232 & 0.5463 \\ \midrule
        \textbf{ADA} & 0.1629 & 0.0142 & 0.1298 & 0.0032 & 0.1556 & 0.0066 & 5.3014 & 0.1821 & 6.8771 & 1.2666 \\ \bottomrule     \toprule
    \end{tabular}
    }
    \bigskip
    \centering
    \resizebox{0.8\textwidth}{!}{
    \begin{tabular}{lcccccccccc}

        & \multicolumn{4}{c}{\textbf{DTI}} & \multicolumn{4}{c}{\textbf{Poverty Map}} \\
        ~ & \multicolumn{2}{c}{avg. R} & \multicolumn{2}{c}{worst R } & \multicolumn{2}{c}{avg. R} & \multicolumn{2}{c}{worst R} \\
        ~ & mean & std & mean & std & mean & std & mean & std \\ \midrule
        \textbf{ERM} & 0.4827 & 0.0080 & 0.4391 & 0.0154 & n/a & n/a & n/a & n/a \\ 
        \textbf{Mixup} & 0.4589 & 0.0131 & 0.4239 & 0.0025 & n/a & n/a & n/a & n/a \\ 
        \textbf{ManiMixup} & 0.4736 & 0.0040 & 0.4306 & 0.0087 & n/a & n/a & n/a & n/a \\ 
        \textbf{Local Mixup}  & 0.4700 & 0.0127 & 0.4325 & 0.0075 & n/a & n/a & n/a & n/a \\ 
        \textbf{CMixup} & 0.4735 & 0.0041 & 0.4346 & 0.0082 & 0.8040 & 0.0396 & 0.5388 & 0.0725 \\ 
        \textbf{ADA} & 0.4928 & 0.0098 & 0.4483 & 0.0094 & 0.7938 & 0.0328 & 0.5218 & 0.0616 \\ 
    \bottomrule
    \end{tabular}
    }
\end{table}

\newpage

\end{document}